%% file: SLIDE_DeepLearningForMBS.tex
\documentclass[11pt,a4paper]{article} 

\usepackage{hyperref}
\usepackage{todonotes} 
\usepackage{xcolor}
\usepackage{amsmath}

\usepackage{acronym} 
\usepackage{mathtools}
\usepackage{siunitx}  
\usepackage{placeins}
\usepackage{a4wide}
\usepackage{doi}

\acrodef{DOF}{degrees of freedom}
\acrodef{LTG}{local to global}
\acrodef{RHS}{right hand side}
\acrodef{ODE2}{second order differential equations}
\acrodef{ODE1}{first order differential equations}
\acrodef{AE}{algebraic equations}
\acrodef{Data}{data variables}

\newcommand{\exuUrl}[2]{\url{#1}} 

\newcommand{\ba}{\begin{array}{c}}
\newcommand{\ea}{\end{array}}

\newcommand{\be}{\begin{equation}}
\newcommand{\ee}{\end{equation}}

\newcommand{\bea}{\begin{eqnarray}}
\newcommand{\eea}{\end{eqnarray}}

\newcommand{\bi}{\begin{itemize} \itemsep -2pt}
\newcommand{\ei}{\end{itemize}}

\newcommand{\bn}{\begin{enumerate} \itemsep -2pt}
\newcommand{\en}{\end{enumerate}}

\newcommand{\eq}[1]{Eq.\ (\ref{#1})}
\newcommand{\eqs}[1]{Eqs.\ (\ref{#1})}

\newcommand{\fig}[1]{Fig.\ \ref{#1}}

\newcommand{\mytab}[1]{Tab.\ \ref{#1}}
\newcommand{\eqdot}{\, .} 
\newcommand{\eqcomma}{\, ,} 

\newcommand{\tp}{^\mathrm{T}}

\newcommand{\bv}{\mathbf{b}}
\newcommand{\ev}{\mathbf{e}}
\newcommand{\fv}{\mathbf{f}}

\newcommand{\qv}{\mathbf{q}}
\newcommand{\rv}{\mathbf{r}}
\newcommand{\pv}{\mathbf{p}}
\newcommand{\tv}{\mathbf{t}}
\newcommand{\uv}{\mathbf{u}}
\newcommand{\vv}{\mathbf{v}}

\newcommand{\xv}{\mathbf{x}}
\newcommand{\yv}{\mathbf{y}}
\newcommand{\zv}{\mathbf z}

\newcommand{\Am}{\mathbf{A}}
\newcommand{\Bm}{\mathbf{B}}
\newcommand{\Dm}{\mathbf{D}}
\newcommand{\Km}{\mathbf{K}}
\newcommand{\Mm}{\mathbf{M}}
\newcommand{\Nm}{\mathbf{N}}

\newcommand{\Rm}{\mathbf{R}}
\newcommand{\Tm}{\mathbf{T}}
\newcommand{\Um}{\mathbf{U}}
\newcommand{\Vm}{\mathbf{V}}
\newcommand{\Wm}{\mathbf{W}}

\newcommand{\tlambda}{\mbox{\boldmath $\lambda$ \unboldmath}\!\!}

\newcommand{\LU}[2]{\prescript{#1}{}{#2}\,}

\newcommand{\LURU}[4]{\prescript{#1}{}{#2}_{#3}^{#4}\,}

\definecolor{deepblack}{rgb}{0,0,0}
\definecolor{lightgrey}{rgb}{0.6,0.6,0.6}
\definecolor{deepblue}{rgb}{0,0,0.8}
\definecolor{deepred}{rgb}{0.6,0,0.4}
\definecolor{warningRed}{rgb}{1,0.2,0.2}
\definecolor{deepgreen}{rgb}{0,0.65,0}
\definecolor{commentgreen}{rgb}{0.007,0.54,0.06}
\definecolor{deepgreen}{rgb}{0,0.5,0}
\definecolor{lightgreen}{rgb}{0,0.75,0}

\usepackage{listings}
\usepackage{amsfonts} 

\usepackage{tikz}

\newcommand{\tikzpicAnnotate}[4]{
{}{}{}
\begin{tikzpicture}
	\node[above right, inner sep=0] (image) at (0,0) {
    \includegraphics[width=\columnwidth]{#1}
};
\node [anchor=south west] (note) at (#3,#4) {#2};
\end{tikzpicture}
}	

\newcommand{\minipageFig}[3]{
\begin{minipage}[c]{#1\columnwidth}
\tikzpicAnnotate{#2}{#3}{-0.5}{-0.5}
\end{minipage}
}

\newcommand{\inp}{\hat{\xv}}

\newcommand{\finp}{\hat{f}}

\newcommand{\outp}{\hat{\yv}}
\graphicspath{{figures/}}

\newcommand{\peter}[1]{{#1}}
\newcommand{\alex}[1]{{#1}}
\newcommand{\joh}[1]{{#1}}

\newcommand{\tin}{t_{\rm in}}
\newcommand{\Tin}{T_{\rm in}}
\newcommand{\tout}{t_{\rm out}}
\newcommand{\Tout}{T_{\rm out}}
\newcommand{\nin}{n_{\rm in}}
\newcommand{\nout}{n_{\rm out}}

\newcommand{\reals}{\mathbb R}

\title{SLIDE: A machine-learning based method for forced dynamic response estimation of multibody systems}

\begin{document}
\maketitle
\begin{center}
\large Peter Manzl$^{1}$, Alexander Humer$^{2}$, Qasim Khadim$^{3}$, Johannes Gerstmayr$^{1}$\\ \vspace{4pt}
1: University of Innsbruck, Austria \\
2: Johannes Kepler University Linz, Austria \\
3: University of Oulu, Finland \\
Correspondence: 
\href{mailto:peter.manzl@uibk.ac.at}{peter.manzl@uibk.ac.at}, 
\href{mailto:alexander.humer@jku.at}{alexander.humer@jku.at}, 
\href{mailto:qasim.khadim@oulu.fi}{qasim.khadim@oulu.fi}, 
\href{mailto:johannes.gerstmayr@uibk.ac.at}{johannes.gerstmayr@uibk.ac.at}
\end{center}

\section*{Abstract}
In computational engineering, enhancing the simulation speed and efficiency is a perpetual goal. 
To fully take advantage of neural network techniques and hardware,
we present the SLiding-window Initially-truncated Dynamic-response Estimator (SLIDE), a deep learning-based method designed to estimate output sequences of \alex{mechanical systems and multibody systems, in particular, which are subject to forced excitation.}
A key advantage of SLIDE is its ability to estimate the dynamic response of damped systems without requiring the full system state to be known, making it particularly effective for flexible multibody systems. 
%
The method truncates the output window based on the decay of initial effects due to damping, which is approximated by the complex eigenvalues of the system's linearized equations. 
%
In addition, a second neural network is trained to provide an error estimation, further enhancing the method's applicability. 
The method is applied to a diverse selection of systems, including the Duffing oscillator, a flexible slider-crank system, and an industrial 6R manipulator, mounted on a flexible socket. 
Our results demonstrate significant speedups from the simulation up to several millions, exceeding real-time performance substantially. 

{
\textbf{keywords}: 
surrogate models, deep neural networks, deep learning, multibody system dynamics, error estimator}
%

\input{01_Introduction.tex}

\input{02_Framework_Methods.tex}
\input{03_MultibodySystems}

\input{05_Conclusion.tex}

\input{Appendix.tex}
\section*{Acknowledgments}
The computational results presented here have been achieved in part using the LEO HPC infrastructure of the University of Innsbruck. 

\section*{Disclosure Statement}
No potential conflict of interest was reported by the author(s).


\bibliographystyle{ieeetr}

\bibliography{bibliographyDoc}

\end{document}

%% file: 01_Introduction.tex
%
\section{Introduction}
\joh{There is no risk of overstatement in saying that neural networks have conquered many, if not all, aspects of life.}
\alex{
Besides the computational power that has come around, a fundamental mathematical property provides the foundation for the -- not only for layman -- astonishing advances, irrespective of whether processing of visual data or large language models are concerned:}
Neural networks \alex{are} universal function approximators~\cite{1989_Hornik_FFNareUniversalFunApproximators} and are applied in very diverse fields \alex{as, e.g.,} image processing~\cite{2012_krizhevsky_AlexnetPaper, 2016_He_deepResidualLearningForImageRecognition}, \alex{reinforcement learning to master} Atari games~\cite{2013_Mnih_PlayingAtari_DQN} \alex{or defeat the world's best humans in} Go~\cite{2016_Silver_masteringGoWithNN}, and natural language processing~\cite{2017_Vaswani_attentionIsAllYouNeed}. 

\alex{
  Naturally, the recent breakthroughs raise the question of whether and how we can harness deep-learning methods and neural networks, in particular, in our very own realm, i.e., computational analysis of multibody systems.
  As a matter of fact, the application of neural networks to engineering problems is by no means a recent topic but dates back many decades, see, e.g., \cite{2022_VuQuocHumer_DeepLearningComputationalMechanics} for an exposition of historic developments.}
\\
Neural networks are applied as black-box models to many different problems, for dynamics and computational engineering some specific methods evolved. 
Rabczuk and Bathe \cite{2023_rabczuk_machineLearningInModelingAndSimulation} give an overview of the different machine learning methods for modeling and simulation, which are not only neural network based. 
In physics-informed neural networks (PINNs)~\cite{2019_Raissi_PINNs} partial differential equations are incorporated in the neural network's loss to avoid learning unphysical solutions. The partial derivatives can be used from the backpropagation algorithm, used for learning the neural network's parameters. 
Hamiltonian neural networks~\cite{2019_Greydanus_hamiltonianNN} are inspired by Hamiltonian \joh{mechanics}, and just like Lagrangian neural networks~\cite{2020_Cranmer_lagrangianNN}
learn \joh{conservation laws and invariances}. 

Multibody system dynamics place special demands on numerical methods, as, when using redundant coordinate formulations, they are often not described by ordinary differential equations (ODEs) or partial differential equations (PDEs) but by differential-algebraic equations (DAEs). The algebraic equations result from constraints on the system such as prismatic joints or rotational joints. 
\joh{Therefore, PINNs, which are commonly applied to problems like fluid mechanics~\cite{2021_Cai_PINNsForFluidDynamics} and heat transfer~\cite{2021_Cai_PINNsHeatTransfer}, can not be directly used because equations of motion are formulated as ODEs or DAEs.} 
Choi et al.~\cite{2021_Choi_dataDrivenSimulation_DeepNerualNetworks} developed an approach for surrogate models of rigid multibody systems, training meta-models of a single and double pendulum, slider-crank, and a transmission system. Han et al.~\cite{2021_Han_DNNFlexible} focuses on a special training algorithm and examines flexible multibody systems, including a piston-cylinder which kinematically corresponds to the slider-crank, and an excavator's flexible boom. 
Pikuli\`{n}ski et al.~\cite{2024_Pikulinski_DataDrivenInverseDynamics} apply neural networks in combination with online error learning for the inverse dynamics of a manipulator with two degrees of freedom and flexure joints. 
For multibody systems it has been shown in~\cite{2021_Angeli_DeepLearningMinimalCoordinates} that neural networks can be used to learn minimal coordinates using the autoencoder structure and applied it to a two-bar mechanism as well as a suspension.  
In ~\cite{2024_Slimak_overviewDesignConsiderationDataDrivenTimeStepping} neural networks are used as a time-stepping scheme, predicting the state vector \joh{in every time-step}, and the solution of the previous step is used autoregressively. 
In~\cite{2024_NajeraFlores_DaneQuinn_StrucurePreservingMLFrameworkIsolatedNonlinearities},  neural networks are applied to isolated subdomains with nonlinearities. 
A different application of artificial intelligence to multibody system dynamics was shown in ~\cite{2024_Gerstmayr_multibodyModelsFromNaturalLanguage}, where natural language is used to create simulation models to assist and speed up the development of the models -- similar to tools like Github Copilot or Visual Studio IntelliCode, which were shown to have a big impact on productivity~\cite{2024_Ziegler_GitHubCopilotImpact}. 

One way to categorize the application to dynamic systems is to divide by how the neural network is embedded into the application. 
Many research papers such as~\cite{2024_Slimak_overviewDesignConsiderationDataDrivenTimeStepping, 2024_NajeraFlores_DaneQuinn_StrucurePreservingMLFrameworkIsolatedNonlinearities, 2024_Han_DataDrivenForcePredictionHydraulics}, and~\cite{2024_WangNegrut_MBDNode} apply the neural network inside a discrete time-step, where the solution for the next time-step is coming either directly from the neural network or time-integration is applied subsequently. This leads to many network passes for a given time sequence. 
\joh{In contrast}, neural networks can also be used to directly predict sequences, as in~\cite{2021_Choi_dataDrivenSimulation_DeepNerualNetworks}, and~\cite{2021_Han_DNNFlexible}.

\joh{The objective of the present paper is to approximate the multibody system's simulation results -- usually only a few measured quantities -- using forward neural networks, thus leading to tremendous speedups when compared to conventional time integration. 
This vast speedup facilitates real-time approximations of vibration, deformation or similar quantities which enable advanced control techniques, accompanied by the simplicity of the neural networks implementation that facilitates implementation on embedded hardware.
}
In contrast to \cite{2024_Slimak_overviewDesignConsiderationDataDrivenTimeStepping}, where the neural network represents the discrete system to predict single steps, our aim is to approximate entire sequences of time-steps, thus avoiding phase shift in the results. 
\joh{Thus, we present SLIDE, a new method for damped systems, predominantly designed for forced or (displacement-)driven excitation. 
Our method's novelty lies in applying a sliding window, where the window's required length is estimated by the eigenvalues of the linearized equation of motion (EOM), representing the decay of initial effects.
The method enables feedforward neural networks (FFN) to be applied directly to sequences of arbitrary length without knowledge of the system's initial conditions or state. This property is particularly advantageous for systems with flexible coordinates, where the full state can hardly be measured in practice.}
Our specific aims distinguish our method from \cite{2021_Choi_dataDrivenSimulation_DeepNerualNetworks} and \cite{2021_Han_DNNFlexible}, who also explored neural networks in multibody system dynamics for approximation of their behavior. 
\joh{Rather than parameterizing mainly autonomous systems with varying masses or geometry, etc. as done in \cite{2021_Choi_dataDrivenSimulation_DeepNerualNetworks}, our approach with driven systems leads to smaller networks and training times, and does not require information on initial values.
Furthermore, we present an error estimator network, which is able to approximate the accuracy of the solution beyond the range where the input-output behavior has been trained. } 

\subsection{Surrogate models for multibody systems} \label{sec:MBD}
\alex{
  Regarding the terminology, the present approach can be attributed to the areas of {surrogate models} and (parametric) {model(-order) reduction}. 
  We refer to the seminal review of Benner et al.~\cite{2015_Benner_survey} and references therein regarding the definition and categorization of surrogate models.
  The aim of surrogate models in dynamic systems and multibody systems, in particular, is to reproduce some specific input-output behavior. 
  In the context of multibody systems, the input-to-output map could, for instance, be some specific load-displacement behavior as, e.g., map from motor torques to the tool center point (TCP) position of a robot. 
}
\alex{
  Surrogate models are closely related to parametric model reduction, i.e., methods to construct reduced order models (ROMs) for which some parameters are not fixed, but may vary within certain, typically pre-defined ranges. 
  Optimization is but one natural application of parametric ROMs.
  The reduction of computational costs is also essential in control tasks and model-predictive control, in particular, and uncertainty quantification~\cite{2015_Benner_survey}.
}
%
%

\begin{figure}%
\includegraphics[width=\columnwidth]{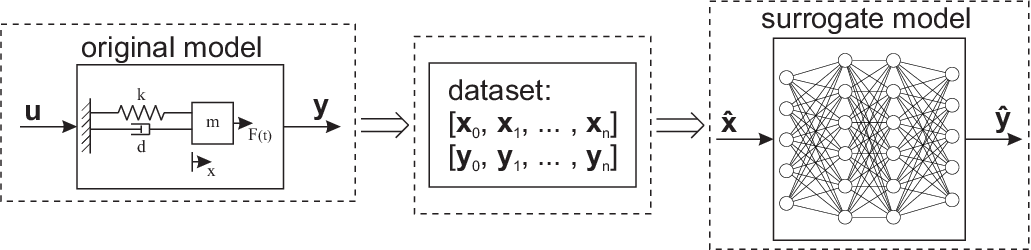}%
\caption{The components of the explored surrogate models: using an original model the dataset is created. The neural network surrogate model is trained and evaluated using this dataset and reproduces the input-output behavior of the system. }%
\label{fig:Concept}%
\end{figure}

The current study focuses on the dynamic response of mechanical and multibody systems. This concept is illustrated in \fig{fig:Concept}, where the surrogate model is constructed from a dataset. This dataset can originate either from measurements, a simulation model, or a combination. %
\alex{
  Mechanical systems are characterized by their nature of being second-order in time. }%
Multibody systems are mechanical systems that consist of interconnected rigid and deformable components, can undergo large translational and rotational displacements~\cite{2020_Shabana_dynamicsofMBS}. 
 Joints that constrain the relative motion of bodies are a defining feature of multibody systems.
\alex{
  Let $\qv \in \reals^{n_q}$ denote the vector of generalized and redundant coordinates that describes a system's current state at position level. Then, the differential-algebraic equations of motion of a holonomic multibody system can be written as
  \begin{align}
    \Mm(\qv) \ddot \qv + \left(  \mathbf{G}_{\qv} \right)\tp \tlambda &= \fv (\qv, \dot \qv, \uv, t) , \label{eq:MBS1} \\
    \mathbf{g} (\qv, t) &= \mathbf 0 , \label{eq:MBS2} \\
    \yv &= \yv(\qv, \dot \qv, \uv, t) \eqdot \label{eq:MBSout} 
  \end{align}
  Here, the vector $\yv \in \reals^{n_y}$ in \eq{eq:MBSout} represents all relevant outputs of the system a user is interested in.
  In the above equations, $\Mm \in \reals^{n_q \times n_q}$ denotes the \alex{(generally deformation-dependent)} mass matrix; the vector of generalized forces $\fv \in \reals^{n_q}$ comprises both internal (elastic) forces, applied forces as well as velocity-dependent, nonlinear inertia forces.
  The redundant coordinates are constrained by algebraic relations $\mathbf{g} \in \reals^{n_a}$, representing position-level holonomic constraints,
  which are enforced by means of Lagrange multipliers $\tlambda \in \reals^{n_a}$ in direction of the constraint Jacobian $\mathbf{G}_{\qv} = \partial \mathbf{g} / \partial \qv$. 
	To fully define the dynamic system, initial conditions need to be specified, 
\be
  \qv(0) = \qv_{|t=0}\eqcomma \quad \mathrm{and} \quad \dot \qv(0) = \dot \qv_{|t=0}\eqcomma 
\ee
	which also must fulfill the constraint equations \eqref{eq:MBS2}. 
  For the sake of brevity, the equations of motion are written only for holonomic constraints, while the overall method can be directly applied to non-holonomic constraints, without general restrictions.
  All our numerical examples are special cases of the general system of equations~\eqref{eq:MBS1}--\eqref{eq:MBS2}.   
}

\joh{We like to mention that the set of DAEs \eqref{eq:MBS1}--\eqref{eq:MBS2} require special implicit time-integration methods, 
where only a few variants can be applied to multibody systems, as compared to the large number of explicit time integration for ODEs.
Moreover, the implicit nature of DAE integrators impedes a potential gain in simulation performance, even for the multi-threaded implementation \cite{2024_Gerstmayr_Exudyn}, which is a further motivation for the present study. }
Note that the differential-algebraic nature of the equations of motion also poses challenges in the application of other neural-network-based approaches such as PINNs~\cite{2019_Raissi_PINNs}. 

\alex{
  Throughout the present paper, we assume constant step-sizes $h$; the displacement vector at the $i$-th time-step is denoted by $\qv_i = \qv_{|t=ih}$.
}

\subsection{Feedforward networks (FFN)}
\begin{figure}[b!]%
\centering
\includegraphics[width=0.75\columnwidth]{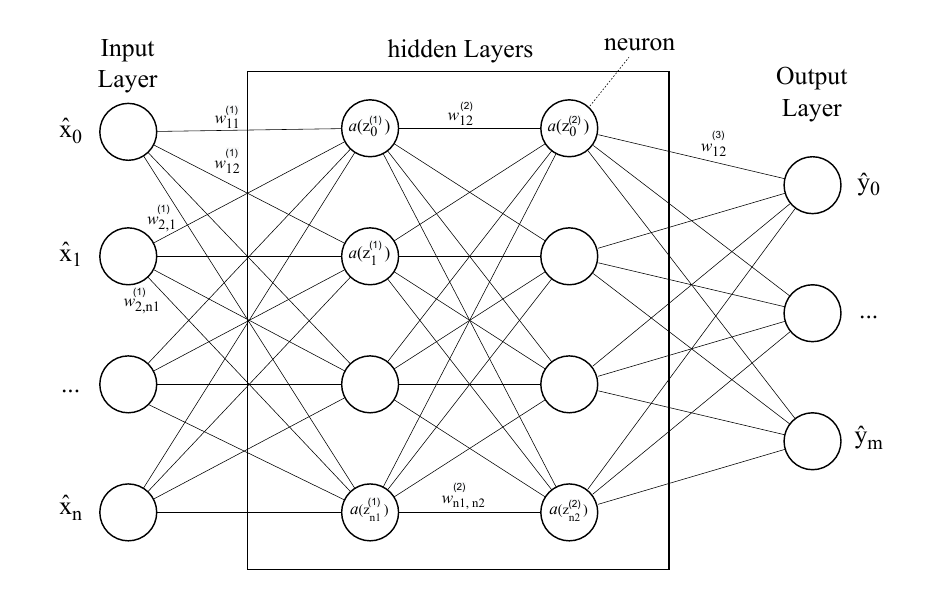}%
\caption{Structure of a general feedforward network. The layers are connected with weights $\Wm$
The hidden layers typically utilize a nonlinear function $a(z)$ \alex{as activation function.}}%
\label{fig:FFNStructure}
\end{figure}
\alex{
  Feedforward neural networks (FFNs), which are also referred to as multi-layer perceptrons (MLPs), are the most fundamental artificial neural networks, which, at the same time, serve as building blocks in many advanced network structures of today's deep-learning methods~\cite[chapter 6]{2016_Goodfellow_DeepLearning}. 
  FFNs are organized in layers of neurons with successive layers being fully connected, i.e., each neuron in the $i$-th layer is connected to each neuron of the \joh{$(i+1)$-th} layer. 
  The first and last layers represent inputs and outputs of the neural network, respectively. 
  Layers in between are referred to as ``hidden'' layers which are characterized by the presence of generally nonlinear activation functions $a$.}
\alex{
  The input $\xv^{(i)}$ to the $i$-th layer, which is $n_i$ neurons wide, is the output of the previous ($i-1$) layer, i.e., $\xv^{(i)} = \yv^{(i-1)} \in \reals^{n^{(i-1)}}$.
  The output of the layer $\yv^{(i)} \in \reals^{n^{(i)}}$ is computed as an affine map $\zv^{(i)}$, to which the activation function $a$ is applied in an element-wise way:  %
  \begin{equation}
    \zv^{(i)} = \Wm^{(i)} \xv^{(i)} + \bv^{(i)} , \qquad
    \yv^{(i)} = a (\zv^{(i)}).
  \end{equation}
}%
\alex{
  The weight matrix $\Wm^{(i)} \in \reals^{n^{(i)} \times n^{(i-1)}}$ represents the connections among the neurons of the $(i-1)$-th layer and the $i$-th layer of the network. The vector $\bv^{(i)} \in \reals^{n^{(i)}}$ represents the biases of the $i$-th layer's neurons. 
  The rectified linear unit (ReLU), the hyperbolic tangent, and the sigmoid function are some of the most commonly used activation functions~\cite{2022_Dubey_ActivationFunctionsSurveyAndBenchmark}.
  The input to the first layer is denoted by $\hat \xv = \xv^{(1)}$; for the output of a network with $L$ hidden layers, we write $\hat \yv = \zv^{(L+1)}$.
  The set of weight matrices $\Wm^{(i)}$ and bias vectors $\bv^{(i)}$ of a network constitute the parameters that are learned when training the network. 
  The number of parameters determines the capacity of a neural network, i.e., its capability to learn complex representations. 
  A network is considered deep if more than one hidden layer is present. 
  Deep networks are generally more difficult to train due to effects like vanishing or exploding gradients, see, e.g.,~\cite{2016_Goodfellow_DeepLearning}.
}%
In \fig{fig:FFNStructure}, a basic neural network with 2 hidden layers is shown exemplary, therefore this network consists of 3 weight matrices and 2 layers with activation functions. 
The input layer $\inp = \left[\hat{x}_0,\; \hat{x}_1,\; ..., \; \hat{x}_{n_\mathrm{in}} \right]$ and output layer $\outp = \left[\hat{y}_0, \; \hat{y}_1 ,\; ..., \; \hat{y}_{n_\mathrm{out}}\right]$ are connected to the first and last layer. 
\joh{
  In the context of this work, we need to distinguish inputs to dynamic systems $\uv$, e.g., generalized forces, prescribed displacements, or rotations, from inputs to neural networks.
  In what follows, the latter may be comprised from the system state $\qv$ and $\dot \qv$, system inputs $\uv$, initial conditions $\xv_{|t=0}$ and $\dot \xv_{|t=0}$, and control inputs (or subsets thereof).
}

In standard feedforward neural networks, the sequence length is fixed, but by sliding the input- and output-windows over a longer sequence and concatenating the outputs, results for longer sequences can be obtained under certain aspects. 


%% file: 02_Framework_Methods.tex
\section{Prediction of dynamic response of multibody systems}\label{sec:methods}

The present section introduces the general concept of the SLIDE method and of the error estimation network. It furthermore shows the computational approach for decay times of initial effects in damped multibody system and shows the calculation of losses during training and validation.
However, the details of implementation will be available open source on GitHub\footnote{\url{https://github.com/peter-manzl/SLIDE}}. 
\subsection{The SLIDE method
}\label{sec:MethodAsymWindow}

\begin{figure}[tb]
\centering
\includegraphics[width=0.7\columnwidth]{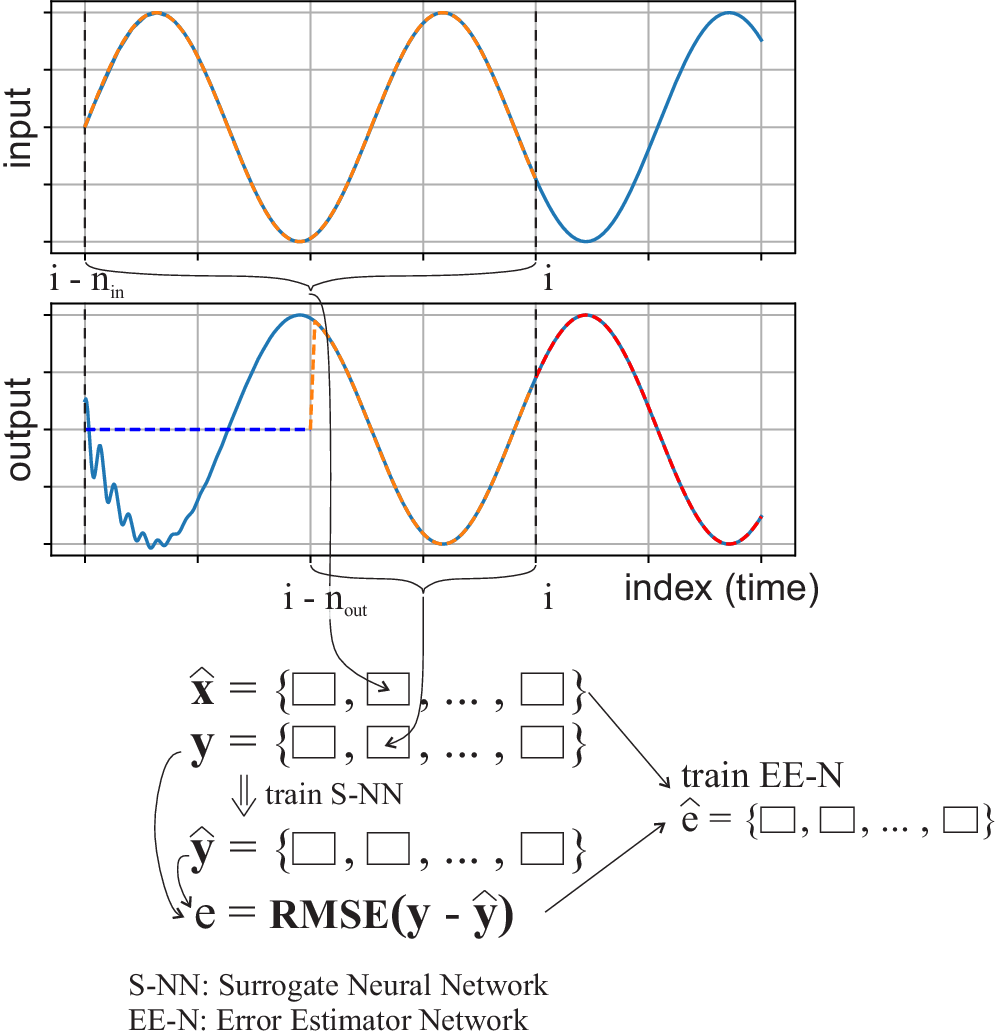}%
\caption{The SLIDE method uses an input window of length $n_\mathrm{in}$ and an output window of length $n_\mathrm{out}$. Because of the damping in the mechanical system, initial conditions and oscillations vanish. The Surrogate Neural Network (S-NN) is trained to map the input $\hat{\xv}$ to the output $\hat{\yv}$. After training the S-NN, the Error Estimator Network (EE-N) is trained to predict the RMSE between the dataset and the neural network's output from the system's input. }%
\label{fig:slideMethod}%
\end{figure}

We propose the \emph{SLiding-window Initially-truncated Dynamic-response Estimator} (SLIDE) method, which is illustrated in \fig{fig:slideMethod}. 
\alex{
  The idea underlying SLIDE is based on the fact that the influence of initial conditions on the evolution of a damped mechanical system decays exponentially over time.
  For linear systems, the influence of the initial conditions is described by the homogeneous solution, i.e., a solution of the homogeneous ODE, which is governed by the eigenvalues of the dynamic system.}
  \joh{Inhomogeneous nonlinear systems generally do not allow for an additive decomposition of the solution into homogeneous and particular parts, and nonlinear damping mechanisms, such as friction, can not be characterized by eigenvalues of a linearized system. } 
  \alex{Still, transient effects related to the initial conditions decay if dissipative mechanisms are present. }
  \joh{Few modern mechatronic or robotic systems exist on ground (as compared to space or sea), where uncontrolled initial conditions show long-term effects.}
\alex{
  Owing to the state-dependent (tangent) stiffness and damping properties of nonlinear systems, decay times generally vary over time.
  }
  
\alex{
  The SLIDE method is constructed to deliberately forgo the transient phase when predicting the response of dynamic systems.
  For this purpose, we train a feedforward neural network to map a (temporal) sequence of inputs onto a shorter sequence of outputs, which, as compared to the input sequence, is truncated by the initial phase in which transient effects may dominate the response.
}%
\alex{
  The input sequence is composed from time-discrete samples of system states and inputs, which are combined into vectors $\rv = \rv(\qv, \dot \qv, \uv, t)$, within an input time window $t_i \in \Tin = [t_{0, \text{in}}, t_{1, \text{in}}]$ of length $\tin = t_1 - t_0$.
  %
  We choose a constant step size $h$ such that the length of the time window translates into an integer-valued number of intervals $\nin = \tin / h$, i.e., $t_i = t_{0,{\rm in}} + i h$, with $i \in \{ 0, 1, \ldots , \nin - 1 \}$.
  The $n^{(0)}$-dimensional input to the neural network, which we subsequently refer to as surrogate neural network (S-NN), is therefore given by
  \begin{equation}
    \inp = \left[ \rv_0\tp, \rv_1\tp, \ldots , \rv_{\nin - 1}\tp \right]\tp \in \reals^{n^{(0)}}.
  \end{equation} 
  Note that the endpoint of the input time window is not included according to our indexing convention. 
  In the simplest (and preferred) case in most our examples, only system inputs as, e.g., external loads, are used, i.e., $\rv_i = \uv_i$. 
  In more complex situations, $\rv$ may additionally depend on $\qv$, $\dot \qv$ or $t$, such as on displacements or forces measured in a real system or on initial conditions $\qv(0)$ and $\dot \qv(0)$.
  }

\alex{
  The output sequence contains the samples of $n_y$-dimensional system outputs $\yv_j = \yv(t_j) \in \reals^{n_y}$ within the output time window $t_j \in \Tout = [t_{0, \text{out}}, t_{1, \text{out}}]$. 
  Input and output windows share the same endpoint $t_{1, \text{out}} = t_{1, \text{in}} = t_1$; 
  the output sequence, however, is initially truncated by at least one (but typcally more) time steps.
  Using the same sampling rate for the outputs as for the inputs, the trunctation implies $\nout \leq \nin$ such that $t_j = t_{0, {\rm out}} + j h = t_1 - (\nout - j) h$, with $j = 1, 2, \ldots, \nout$. 
  Accordingly, the output is given by
  \begin{equation}
    \outp = \left[ \yv_1\tp, \yv_2\tp, \ldots , \yv_{\nout}\tp \right]\tp \in \reals^{n^{(L+1)}} , \qquad n^{(L+1)} = n_y \nout ,
  \end{equation}
  where we exclude the initial point of the output time window.
  Therefore, it is guaranteed that input and output sequence are shifted by one time step even in the limiting case of equal window lengths, i.e., $\Tin = \Tout$.
}%

\alex{
  In what follows, we denote the representative time constant, which governs the ``slowest'' relevant decay of initial conditions and perturbations, as $t_d$.
  Accordingly, the length of input and output sequences are supposed to comply with 
  \be \label{eq_truncated_time}
  t_d < t_{0, {\rm out}} - t_{0, {\rm in}} = \left( \nin - \nout \right) h .
  \ee 
}%
\alex{
  The above description uses single input and output windows, respectively. 
  The SLIDE method does not necessarily condition the S-NN on initial conditions, or, more generally, on initial states at the beginning of the input time window.
  In the most simple case, only system inputs are used in the input sequence.
  In this case, we can easly construct a sliding-window approach.
  By introducing a time-shift (stride) of $\tout$ ($\nout$) for both input and output windows, we can predict outputs arbitrarily ahead in time.
}

In this study, the S-NN is trained on input-output sequences without sliding-windows, as the training is not affected by the sliding. 
Data is obtained from simulation and the sliding-windows are applied in testing. 
As in real systems the initial conditions may not be known or only hard to obtain for all states, the SLIDE method enables good predictions without measuring the full state. While the SLIDE method is designed for damped multibody systems, it can in a similar way be applied to systems without damping when the full initial conditions are known for each sequence. 
Then, no truncation of the output window is necessary. 

\subsection{Error Estimation Network}
\begin{figure}[tb]
\centering
\centering
\includegraphics[scale=1]{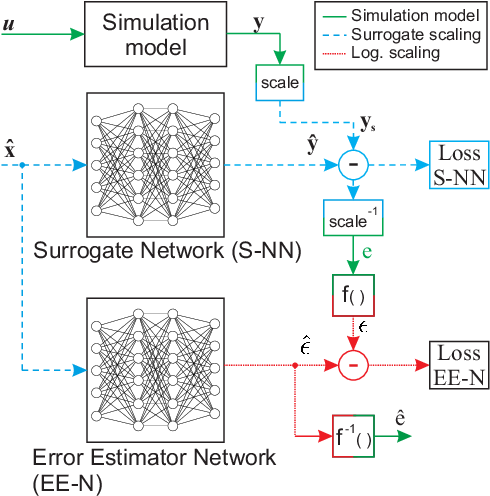}%
\caption{The structure of the proposed error estimator. 
\peter{For better training performance, the output from the simulation model is scaled to normalize $\yv_s$. Before calculating the error of the S-NN, the output is rescaled to its original units. }
The root mean squared error $e$ is transformed to $\epsilon$ by a logarithmic function \eq{eq:logmapping} to \eqref{eq:logmapping2}, denoted by $f()$. The inverse mapping $f^{-1}()$ is used to obtain $\hat{e}$ from the error estimator. 
}
\label{fig:StructureEstimator}%
\end{figure}

It is well-known that \alex{feedforward} neural networks generally
do not extrapolate nonlinear functions well~\cite{2020_Xu_HowNeuralNetworksExtrapolate}. 
\alex{To apply the proposed} neural network surrogate model to real systems, 
\alex{
  we want to have an estimate for the error of the network's prediction similar to the error estimator in time integration methods.
}
\alex{
  For this purpose, we propose to train a second feedforward neural network, which is referred to as Error Estimator Network (EE-N), to provide error estimates for the prediction of the S-NN.
}%
The structure of the EE-N is shown in \fig{fig:StructureEstimator}. 
\alex{
  The error estimator is trained independently from the surrogate model, i.e., parameters of the S-NN are fixed when training the EE-N.
  Additional data is required for the training process, since errors predicted by the S-NN for the training data the S-NN was trained on are (ideally) very small.
  For this reason, additional data outside the previously trained ranges \alex{of inputs and outputs is included, such that} the surrogate model \alex{is forced} to also extrapolate, which results in significantly large errors of the S-NN's predictions.
}%

The target error is the root mean squared error (RMSE) of the difference between true outputs of the multibody system and corresponding predictions by the S-NN  (averaged over an output time window) 
\be\label{eq:errors_estimator}
e = \sqrt{\frac{1}{n^{(L+1)}} \sum_{i=1}^{n^{(L+1)}} \left(\outp_i - \yv_{s, i} \right)} ,
\ee
with the length $n^{(L+1)}$ of the output vector of the S-NN. 
The estimator network is then trained on the logarithmic error $\epsilon$
\be
\epsilon = \frac{\log_{10}(e) - \epsilon_r}{\epsilon_r} \eqcomma \label{eq:logmapping}
\ee
\alex{which is shifted and normalized by}%
\be
\epsilon_r = \frac{\epsilon_+ - \epsilon_-}{2} \label{eq:logmapping2}
\ee
to logarithmically scale errors between $10^{\epsilon_-}$ and $10^{\epsilon_+}$ into the range $[-1, 1]$. The network estimates the logarithmic error $\hat{\epsilon}$, from which the estimated RMSE can be calculated by
\be
\hat{e} = 10^{\left(\epsilon_r \hat{\epsilon} - \epsilon_r\right)}\eqdot
\ee
As the main interest is in the relative accuracy of the solution, this metric helps us achieve more accurate estimates of the error both on \alex{the data the S-NN was trained on and beyond where the error may increase by orders of magnitude.}%
The loss function of the EE-N is again the mean squared error (MSE), see \eq{eq:MSE}, between the estimated logarithmic error $\hat{\epsilon}$ and the target logarithmic error $\epsilon$.

While the EE-N is applied to the input of the surrogate network, it could also be applied to the output or a combination of both. By avoiding using the output of the {S-NN}, both networks can be parallelized for better performance. 


\input{02_1_DampingInMBS}

%
\subsection{Training loss and validation error}
The neural networks are trained using the mean squared error (MSE) loss
\be
L_\mathrm{MSE} = \frac{1}{n} \sum_{i=1}^{n}\left(\hat{\yv} - \yv_s \right)^2
\ee
as objective function, where $\yv_s$ and $\outp$ denote the (scaled) target output from the dataset and output of the neural network, respectively. 
The neural network's parameters $\Wm$ and $\bv$ are initialized using Xavier uniform distribution~\cite{2010_Glorot_UnderstandingDifficultiesTrainingFFN}. 
In all experiments, the ADAM optimizer~\cite{2014_Kingma_AdamOptimizer} is used. For both the error estimator and the validation, the root mean squared error (RMSE) 
\be
L_\mathrm{RMSE} = \sqrt{\frac{1}{n}\sum_{i=1}^{n}\left(\hat{\yv} - \yv_s \right)^2} \label{eq:MSE}
\ee
is evaluated, since the RMSE is in the magnitude of the occurring errors. 
As customary in deep learning, training proceeds in epochs, where the whole training dataset is traversed once per epoch. 
For a dataset size of $n_d$ and a batch size of $n_b$, ${n_d}/{n_b}$ optimizer steps (\textit{iterations}) are performed in each epoch. 
The batch size describes how many entries of the dataset are passed simultaneously in each iteration. In the later shown examples, the hardware utilization increases with the batch size, in addition, smaller batch sizes also require generally smaller learning rates, as more parameter updates are performed per epoch. 
The dataset is shuffled in every epoch to add stochasticity. 
During training, the validation error is tracked and the weights and biases of the episode with the lowest validation are saved for testing to avoid overfitting.

%% file: 02_1_DampingInMBS.tex
\subsection{Computation of decay time in linearized multibody systems} \label{sec:DampingMBS}

The proposed SLIDE method is based on the assumption that effects due to unknown initial conditions or disturbances \alex{decay within} a certain time, which is the length $t_d$ of the truncated part of the response window, compare \eq{eq_truncated_time}.
A reasonable estimation of $t_d$ can be based on complex eigenvalues of the linearized system, which hereafter is used to compute worse-case scenarios for the damping times of all eigenvalues. 
The computation of eigenvalues for the investigated multibody systems as well as a reasonable guess for $t_d$ are shown in the following.

Using \eqs{eq:MBS1} and \eqref{eq:MBS2}, we linearize the system equations about $\qv_L$, $\dot \qv_L$, and $\tlambda_L$ in the following way,
  \begin{align} \label{eq_linearized1}
    \Mm(\qv_L) \ddot \qv_L + \left(  {{\mathbf{G}}_{\qv_L}} \right)\tp \tlambda_L &= 
      \left. \frac{\partial \fv (\qv, \dot \qv, \uv, t)}{\partial \qv} \right|_{\qv_L, \dot \qv_L} + 
      \left. \frac{\partial \fv (\qv, \dot \qv, \uv, t)}{\partial \dot \qv} \right|_{\qv_L, \dot \qv_L}
      , \\
    \left. \frac{\partial \mathbf{g} (\qv, t)}{\partial \qv} \right|_{\qv_L} \qv_L &= \mathbf 0 . 
  \end{align}
Here, we neglect terms $\partial \Mm(\qv) / \partial \qv$, assuming the linearization evaluated for $\ddot \qv = \mathbf 0$, and we furthermore neglect terms $\partial \left( \left( {\mathbf{G}}_{\qv} \right)\tp \tlambda \right) / \partial \qv$.

The linearized equations are thus transformed into matrix form, where, 
for the computation of eigenvalues, we formulate constraints on the acceleration level:
\be \label{eq_linearized2}
  \begin{bmatrix}
      \Mm(\qv_L) & \left(  {\mathbf{G}}_{\qv_L} \right)\tp\\
      {\mathbf{G}}_{\qv_L} & \mathbf 0\\    
  \end{bmatrix}
  \begin{bmatrix}
      \ddot \qv_L \\
      \tlambda_L  \\
  \end{bmatrix} + 
  \begin{bmatrix} \Dm & \mathbf 0 \\ \mathbf 0 & \mathbf 0 \end{bmatrix} \begin{bmatrix} \dot \qv_L \\ \tlambda_L \end{bmatrix} + 
  \begin{bmatrix} \Km & \mathbf 0 \\ \mathbf 0 & \mathbf 0 \end{bmatrix} \begin{bmatrix} \qv_L \\ \tlambda_L \end{bmatrix} 
  = \begin{bmatrix} \Rm_{\qv} \\ \Rm_{\tlambda}  \end{bmatrix} .
\ee
In the above system of equations, $\Dm$ and $\Km$ denote tangent damping and stiffness matrices, respectively.
We assume both residual forces and external loads to vanish when determining the eigenvalues of the multibody system, i.e., we consider the homogeneous system of ODEs subsequently.

In order to compute the complex eigenvalues of the constrained system, we project the equations into the nullspace of the linearized constraints.
For this purpose, we compute the nullspace of the constraint matrix using the singular value decomposition (SVD) for ${\mathbf{G}}_{\qv_L}$, 
\be
  {\mathbf{G}}_{\qv_L} = \Um_S \boldsymbol \Sigma_S \Vm_S^*
\ee
resulting in the unitary matrices $\Um_S$ with left singular vectors as columns, and $\Vm_S$ with right singular vectors as rows. 
The diagonal matrix $\boldsymbol \Sigma_S$ contains the singular values of ${\mathbf{G}}_{\qv_L}$.

Using a tolerance $s_\mathrm{tol}$ for determining non-zero singular values, we compute the number $n_\mathrm{nz}$ of relevant vectors in $\Um_S$ representing the nullspace $\Nm \in \reals^{(n_\mathrm{q}-n_\mathrm{nz}) \times n_\mathrm{q}}$, with components 
\be
  N_{j,i} = U_{i, j+n_\mathrm{nz}}
\ee
for singular values sorted in descending order in the diagonal of $\boldsymbol \Sigma_S$, thus the columns of $\Um$ are the rows of the nullspace. 
%
The system quantities can be computed conveniently by projection, reading as
\be
  \overline{\Mm} = \Nm \Mm \Nm^T, \quad 
  \overline{\Km} = \Nm \Km \Nm^T, \quad \mathrm{and} \quad
  \overline{\Dm} = \Nm \Dm \Nm^T ,
\ee 
which now represent the linearized system in the constrained space,
\be \label{eq_linearized_projected_EOM}
  \overline{\Mm} \, \ddot{\overline{\qv}}_L + \overline{\Dm} \, \dot{\overline{\qv}}_L + \overline{\Km} \overline{\qv}_L = \overline{\Rm}_L \eqdot
\ee
In order to compute the complex eigenvalues, we transfer \eq{eq_linearized_projected_EOM} into the set of first order differential equations, setting $\overline{\Rm}_L=\mathbf 0$,

\be
  \Am \zv	+ \Bm \dot{\zv}	= 0 \quad \mathrm{with} \quad \zv = \begin{bmatrix}\overline{\qv}_L \\ \dot{\overline{\qv}}_L \end{bmatrix}
\ee
and the matrices
\be
  \Am = \begin{bmatrix} \overline{\Km} & \overline{\Dm} \\ \mathbf 0      &-\overline{\Mm} \end{bmatrix}, \quad \mathrm{and} \quad
  \Bm = \begin{bmatrix} \mathbf 0      & \overline{\Mm} \\ \overline{\Mm} & \mathbf 0      \end{bmatrix}
\ee
and compute the complex eigenvalues $\vv = [v_0, \ldots \;, v_1, \ldots ]$ and the eigenvector matrix ${\mathbf{\Phi}}$ of the system matrix $\Am_\mathrm{sys} = -\Am^{-1} \Bm$.
There are several possibilities in computing $\Am_\mathrm{sys}$ or its variants, however, the latter one being directly related to the according eigenvectors for displacements and velocities, as well, which could be back-projected to unconstrained space of $\qv_L$ and $\dot \qv_L$. 

Based on the system's eigenvalues, the required length of the initially-truncated window for the SLIDE method can be computed.
The linearized system response in the time domain can be computed for certain initial conditions $\qv_0$, which include initial displacements and velocities, as follows, 
\begin{equation}
    {\qv_L}(t) = \sum_{i=1}^{2n_f} {\mathbf{\Phi}}_{i}{\pv_{0,i}} e^{v_i t} 
\end{equation}
where $\pv_{0}$ represents the vector of initial modal coordinates, which can be computed as $\pv_0 = \mathbf{\Phi}^{-1} {\qv_L}_0 $ using the eigenvector matrix ${\mathbf{\Phi}}$.

Pairs of complex conjugated eigenvalues represent underdamped behavior, while purely real eigenvalues show overdamped behavior of underlying coordinates. 
For the linearized system, considering a single eigenvalue $v_i$, the envelope of the relative amplitude 
can be calculated as 
\be
A_\mathrm{rel}(t) = e^{\mathrm{Re}(v_i) t}= e^{-\omega_0 D t} \label{eq:decay}
\ee
from the homogeneous solution with the dimensionless damping ratio $D$ and natural frequency $\omega_0$. The initial conditions are damped down to an $A_{\mathrm{rel}, 1\%}$ after 
\be
t_d = \frac{\log(A_{\mathrm{rel}, 1\%})}{\mathrm{Re}(v)} \eqcomma
\label{eq:dampingTimes}
\ee
Therefore for a spring-damper with $\omega_0 = 40$ and $D=0.1$, the initial conditions are 
damped down to $1\%$ after a time of $t_d = \SI{1.15}{\second}$, which agrees well with the numerical solution. 

As for general multibody systems are nonlinear, the damping can depend on the current configuration. Therefore for analyzing it, the system is randomly initialized in the relevant range several times and the eigenvalues are recorded while the system is in motion.

%% file: 03_MultibodySystems.tex
\FloatBarrier
\section{Application to multibody systems}
\alex{
    In what follows, we apply the proposed SLIDE method to several examples problems. 
    A single-DOF system is meant to illustrate the fundamental ingredients of the approach.
    Subsequently, we conduct experiments on actual multibody systems, i.e., a planar slider-crank model and a flexible robotic system.  
}
Previously the rigid slider-crank system's kinematics was considered i.a. by Choi et al. \cite{2021_Choi_dataDrivenSimulation_DeepNerualNetworks} and Wang et al.~\cite{2024_WangNegrut_MBDNode}, whereas the flexible slider-crank system is investigated by Han et.al~\cite{2021_Han_DNNFlexible} using the floating frame of reference formulation (FFRF). 
Furthermore, a serial robot with $6$ rotational degrees of freedom, standing on a soft flexible socket, is investigated. The robot moves along trajectories with prescribed point-to-point (PTP) motions, which causes the socket to deform, resulting in positioning errors.  
Training, validation, and test data for the supervised learning is created by simulation, which is realized using the open-source multibody dynamics code Exudyn~\cite{2024_Gerstmayr_Exudyn}, which provides a Python interface for its efficient C++ implementation\footnote{Version 1.8, \url{https://github.com/jgerstmayr/EXUDYN}}.
The machine learning library PyTorch~\cite{2024_Ansel_Pytorch2} is used for the creation, training, and application of the shown neural networks\footnote{Version 2.3, \url{https://github.com/pytorch/pytorch}}. 
  \joh{In addition to the network's structure (e.g., depth, width, and activation),  parameters governing the training process (e.g., learning rate, batch size, optimizer-specific parameters) constitute hyperparameters that greatly influence the convergence and performance of the network, which are provided in Appendix \ref{sec:Appendix}.}




\subsection{Introductionary example: the mass spring damper}\label{sec:springDamper}
\begin{figure}[tb]
\centering
\begin{minipage}{0.3\textwidth}
\includegraphics[width=\textwidth]{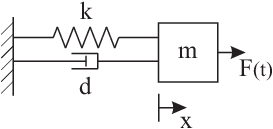} 
\end{minipage}\hspace{1cm}
\begin{minipage}{0.29\textwidth}
\begin{tabular}{c | c}
Parameter & value \\ \hline
$m$ & $\SI{1}{\kilogram}$ \\
$k$ & $\SI{1600}{\N\per\metre}$ \\
$d$ & $\SI{8}{\N\per\metre\per\second}$ \\
$\omega_0$ & $\sqrt{\frac{k}{m}} = 40\si{\per\second}$ \\
$D$ &  $\frac{d}{2 m \omega_0} = 0.1$ \\ \hline \hline
\multicolumn{2}{c}{Duffing oscillator:}\\
$\alpha$ & 0.5
\end{tabular}	
\end{minipage}
\caption{The spring damper model consists of the mass $m$, stiffness $k$ and damping $d$. The natural frequency $\omega_0$ and the dimensionless damping $D$ are derived from these parameters. 
The factor $\alpha$ is used to describe the nonlinearity of the Duffing oscillator. 
}
\label{fig:modelsSpringDamper}%
\end{figure}
The linear mass-spring-damper system, shown in \fig{fig:modelsSpringDamper}, is described by the \alex{following scalar-valued ODE:}
\be
m \ddot{x} + d \dot{x} + k x = F(t)\eqcomma
\ee
where $m$ denotes the mass; $d$ and $k$ are the (viscous) damping parameter and the spring stiffness, respectively. 
\begin{figure}[bt]
\minipageFig{0.54}{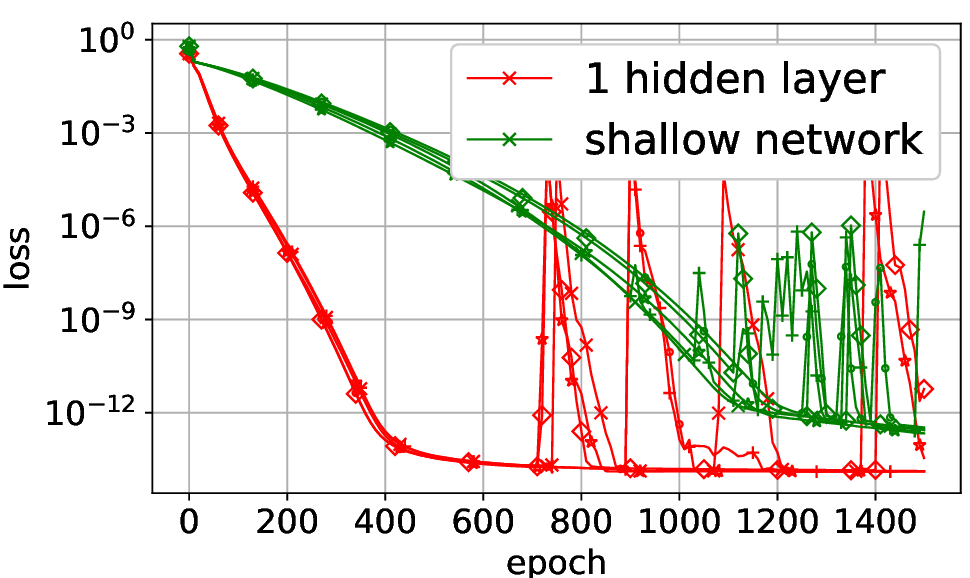}{a)}
\minipageFig{0.44}{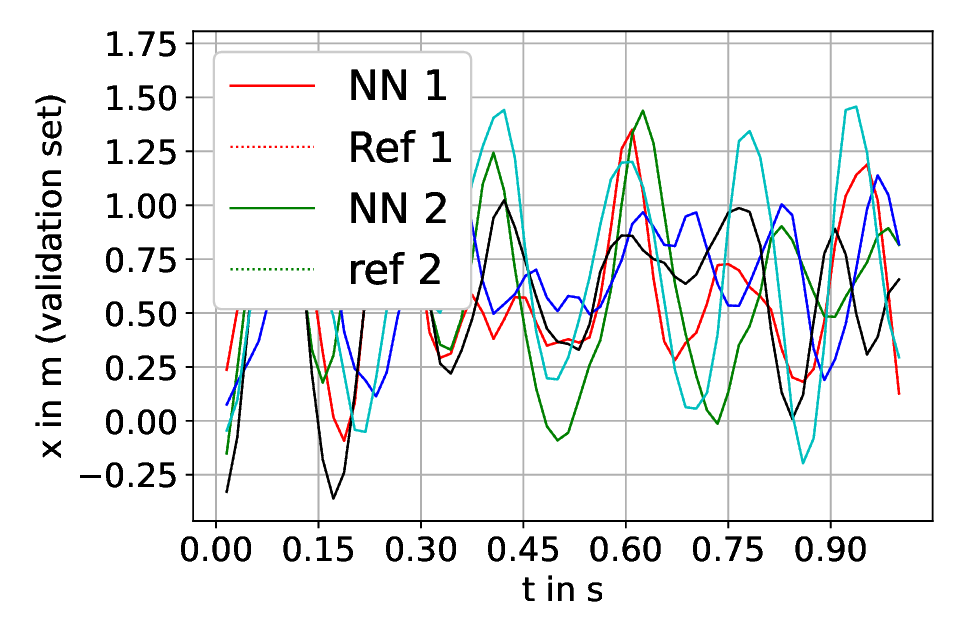}{b)}
\caption{a) The mean square error on the test set while training shown over the epochs. Lines in the same color represent different seeds in the training. b) The results for the validation sets 0 to 4. As the validation error is in the magnitude of $10^{-12}$ to $10^{-14}$, the neuronal network (NN) and reference (time integration) fit almost perfectly numerically.} %
\label{fig:linearTraining}%
\end{figure} 
    The excitation is realized as an external force $F(t)$, which is scaled for training purposes, i.e., $\hat f(t) = \num{5e-4} F(t)$.
    The input vector to the neural network comprises positions $x_0 = x(0)$ and velocities $\dot x_0 = \dot x(0)$ at the beginning of the (single) input time window
    as well as forces $\hat{f}_i$ at the discrete time-steps $t_i = i h$, with $i = 0, \ldots, n_{\rm in} - 1$. 
    In the present example, a window of length $\qty{1}{\second}$, which is discretized into $n_{\rm in} = 64$ steps (step size $h = \qty{15.625}{\milli \second}$), is chosen. 
    The input vector of the first time window therefore reads
    \begin{equation}
        \inp = \left[x_0, \dot{x}_0, \finp_0 \, ... \, ,\,  \finp_{\nin - 1}\right]\tp .
    \end{equation}

The surrogate network is trained to predict the position of the mass, where the output window equals the input time window, i.e., $\nin = \nout$: 
    \begin{equation}
        \outp = \left[ x_1, x_2, \ldots , x_{\nout} \right]\tp ;
    \end{equation}
    inputs and outputs are therefore offset by one time step.
    As opposed to forces, position and velocity are not scaled -- neither in inputs, nor in outputs.
%
No further truncation of the output window is required, since initial conditions $x_0$ and $\dot{x}_0$ are provided as inputs to the network, so also the transient phase can be captured accuratly. 
The simulation does not necessarily use the same stepsize as used for the surrogate model.
In the simulation, the force is constant between the neural network time steps. The last input force $\hat{f}_{n_\mathrm{in}-1}$ is acting until the end of the time-sequence. 
To construct training and test data, we sample from a uniform distribution for all inputs. 
Initial conditions $x_0$ and $\dot{x}$ are set at the beginning of each simulation and in each time-step a random force $F_i \in \left[-2, 2\right] \si{\kilo\newton}$ is applied to the mechanical system.

To represent this linear system, the neural network does not require any nonlinear activation function. 
A neural network with single hidden layer using the identity function as activation and zero biases can be described by
\be
\outp = \underbrace{\Wm^{(2)} \Wm^{(1)}}_{\Wm'} \inp \eqdot
\label{eq:NN1LayerLinear}
\ee
Depending on the exact training parameters, the neural network is able to learn the exact solution of the system with a mean squared error (MSE) on the validation set in the order of $10^{-14}$, as shown in \fig{fig:linearTraining}. The \textit{shallow} network only has one weight matrix connecting input and output, while the other has one hidden layer, described by \eq{eq:NN1LayerLinear}. 

\begin{figure}[tb]
\centering
\minipageFig{0.54}{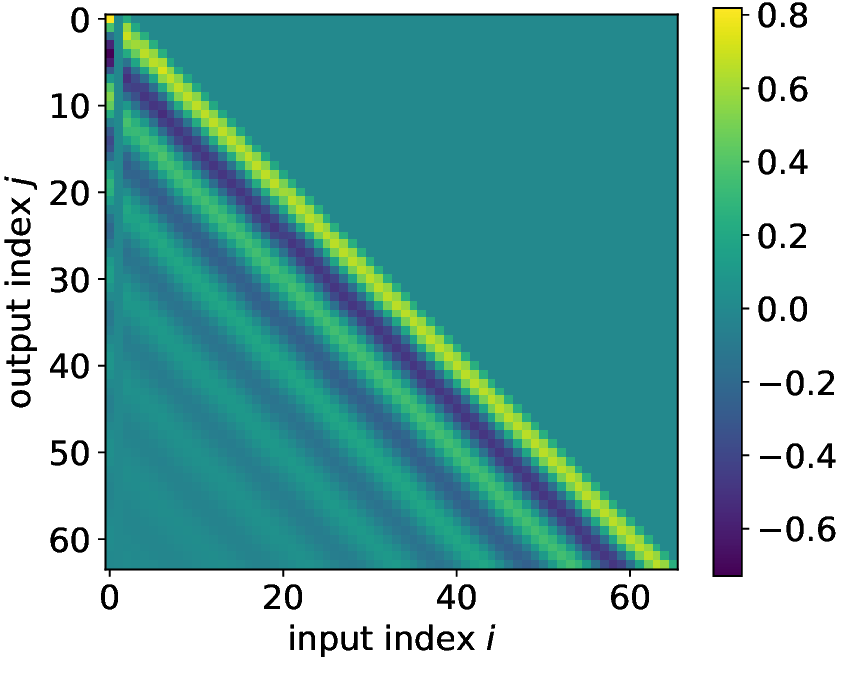}{a)}
\minipageFig{0.43}{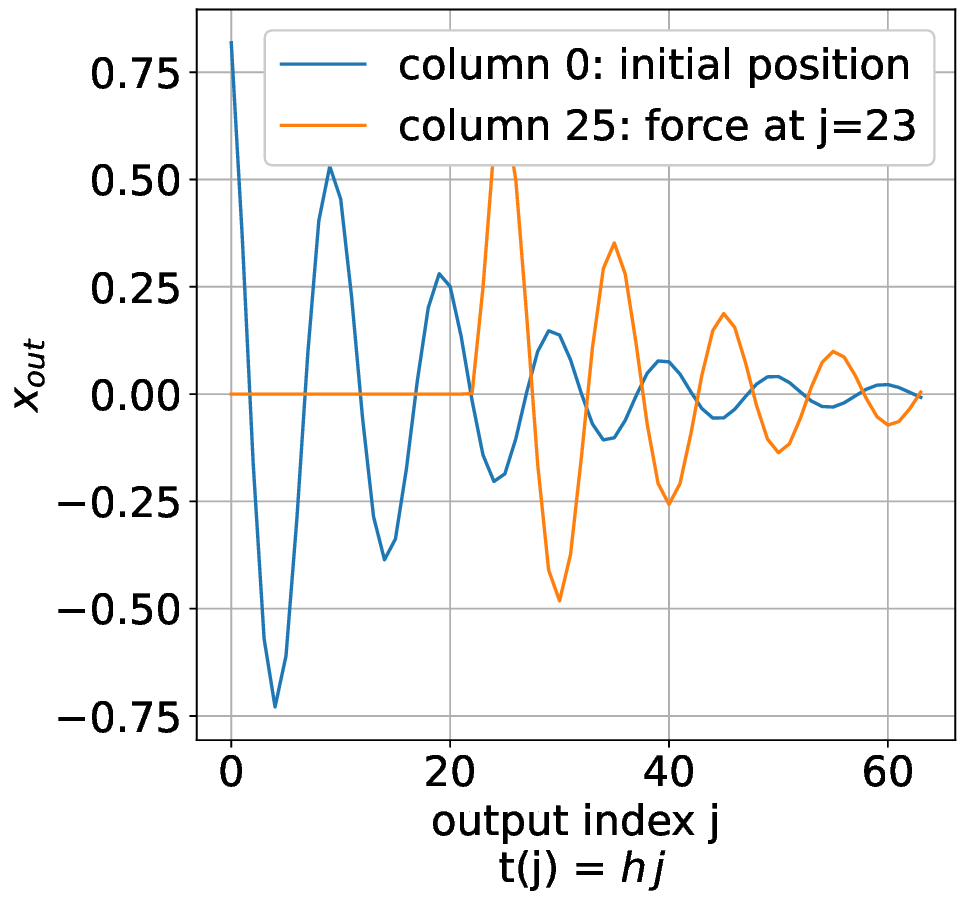}{b)}
\caption{Visualizing of the weight matrix $\Wm'$ as a heatmap. 
The first column shows the homogeneous solution for $x_0 = 1$, which decays over time, represented by the output index. The solution for an arbitrary input vector can be calculated by superimposing the individual solutions - which is what the network's matrix multiplication $\outp = \Wm' \inp$ does.}
\label{fig:linearSpringDamperMapping}%
\end{figure}

The weight matrix $\Wm'$ can be visualized as a heatmap, shown in \fig{fig:linearSpringDamperMapping}. 
The upper right triangle consists of zeros, indicating causality: later inputs do not affect earlier outputs. The first two columns represent the decaying initial conditions over time, while the others represent the system's response to a force impulse at a specific time-step. For predicting the entries at the end of the output, the initial conditions are no longer required. 
Although the neural network was only trained using randomized force input, it is able to process arbitrary input signals with the same discretization because of the known superposition property of ordinary differential equations. 

\subsection{Duffing oscillator: nonlinear spring-damper system}\label{sec:DuffingOsc}
As second example problem, we consider the Duffing oscillator, which is governed by the nonlinear ODE
\be
m \ddot{x} + d \dot{x} + k x + \alpha k x^3= F(t)\eqdot 
\ee
\alex{
    We study the case of a hardening (progressive) spring, $\alpha> 0$.
    As with the linear oscillator, the excitation is scaled for training purposes, i.e., $\hat f(t) = \num{1e-3} F(t)$.
    The natural frequency $\omega_0$ of the undamped Duffing oscillator depends on the displacement $\bar x$, at which the system is linearized, i.e.,
    \begin{equation}
        \omega_0 = \sqrt{\frac{k \left( 1 + 3 \alpha \bar x^2 \right)}{m}} \eqcomma
    \end{equation}
    For the (weakly) damped system, we find a pair of complex conjugate eigenvalues $v_{1,2}$, with a real part of 
    \begin{equation}
        \mathrm{Re} ( v_{1,2} ) = - \omega_0 D = - \frac{d}{2m} \eqcomma 
    \end{equation}
    where the nondimensional damping $D = d / 2 m \omega_0$ has been introduced.
    The real part, whose negative reciprocal is the time constant of the exponential decay of perturbations from the equilibrium state, is independent of the displacement $\bar x$, at which the system is linearized.
}%

\begin{figure}[tb]
\centering
\minipageFig{0.44}{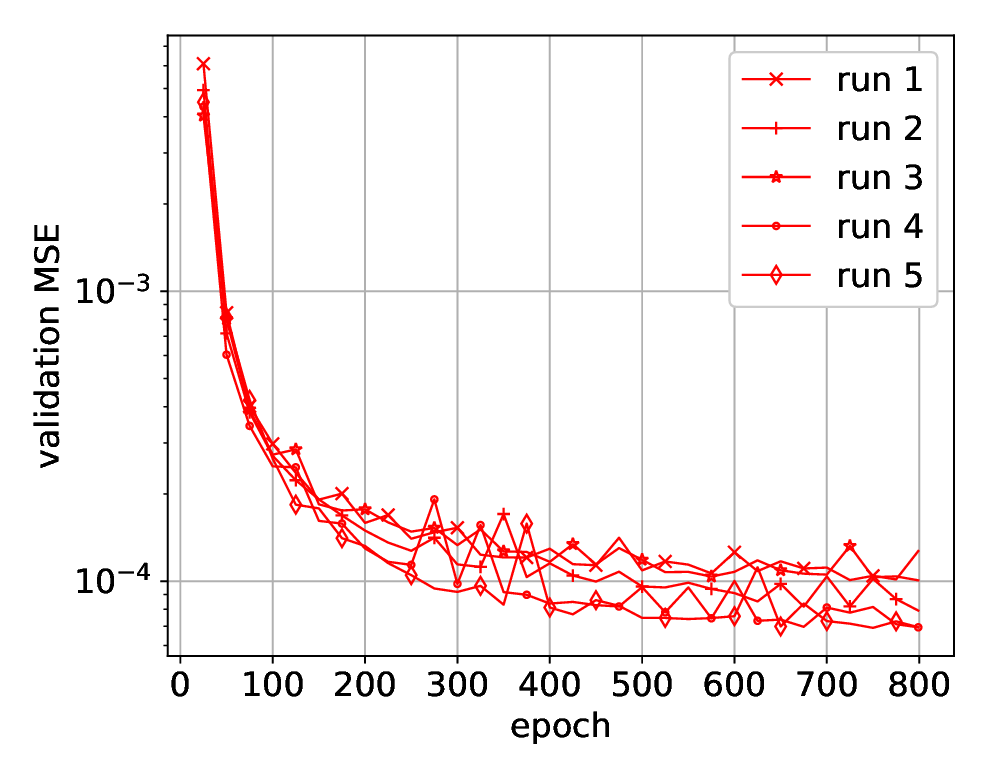}{a)}
\minipageFig{0.44}{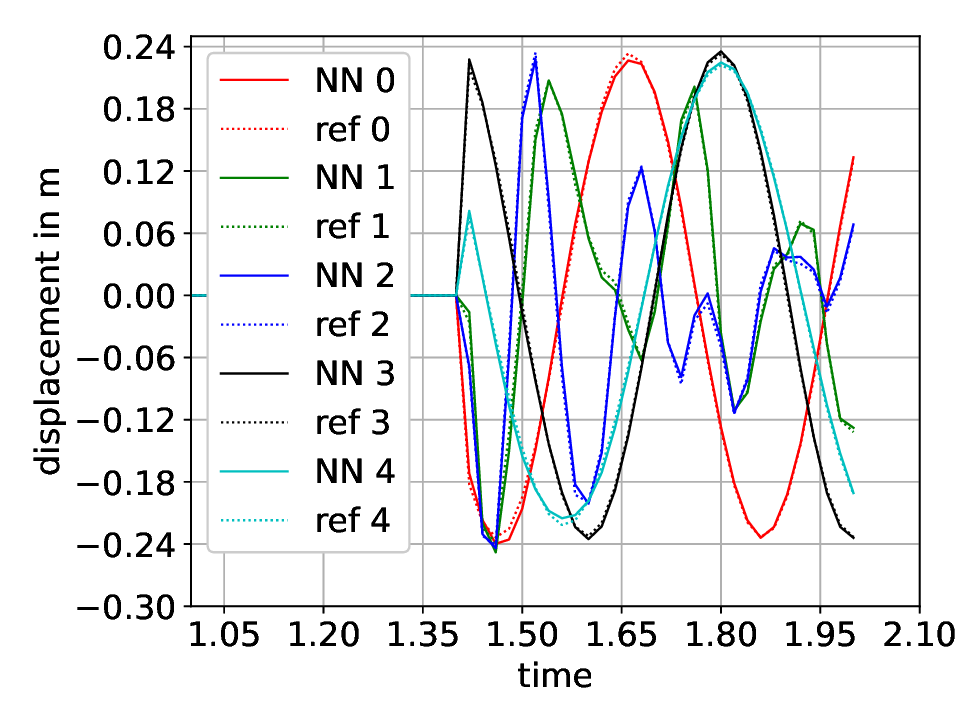}{b)}
\caption{a) The MSE on the validation set of the nonlinear damper over the course of the training using the asymmetric window approach. b) The displacement given by the network of run 1 on the validation set. After training the maximum MSE is $0.19 10^{-3}$ on the training set and $5.65 10^{-3}$ on the validation set. 
}%
\label{fig:DuffingResults}%
\end{figure}

\begin{figure}[tb!]
\includegraphics[width=\columnwidth, trim={0 0 0 1.4cm},clip]{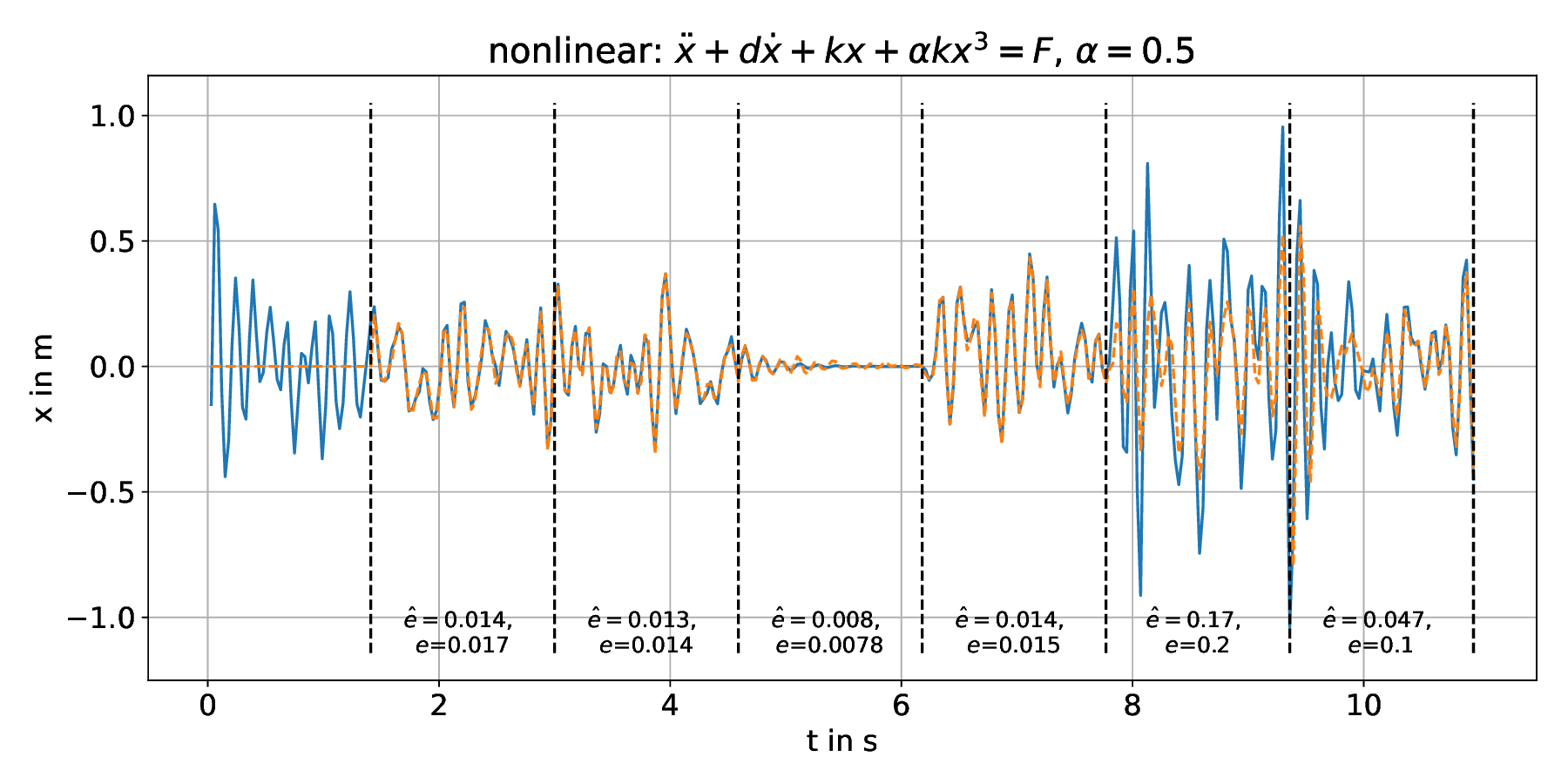}%
\caption{SLIDE applied to a longer input sequence of the nonlinear spring-damper system with parameters from \fig{fig:modelsSpringDamper}. The estimated error $\hat{e}$ is calculated for each segment with the real error $e$ shown below. }%
\label{fig:OsciRecursivelyEst}%
\end{figure}

Thus, for a relative amplitude decay of $1\%$ the time constant $t_d = \SI{1.15}{\second}$ is calculated according to \eq{eq:dampingTimes}. 
The previously described error estimator is applied to the surrogate model. The validation MSE is shown in \fig{fig:DuffingResults}a). For the surrogate neural network (S-NN), 5 trainings are performed, where each time the seed for the initialization of the network weights and the dataset shuffle are initialized differently. For run 1, the results for the first five training sets are shown in \fig{fig:DuffingResults}b). It is also evident that, for the validation data, the network starts the prediction at $t=(n_\mathrm{in} - n_\mathrm{out})h = \SI{1.4}{\second}$, in the training no sliding of the windows is performed. For both training and validation, the input 
\be
\inp = \left[\finp_0,\, \finp_1,\, ...\,,\, \finp_{\nin-1} \right] ,
\ee
is used to obtain the output 
\be
\hat{\yv} = \left[x_{n_\mathrm{in} - n_\mathrm{out}}, x_{n_\mathrm{in} - n_\mathrm{out} + 1},\, ...\,,\, x_{n_\mathrm{in}} \right] \eqdot
\ee
In testing of the trained surrogate model, longer sequences
\be
\inp' = \left[\finp_0,\, \finp_1,\, ...\,,\, \finp_{n_\mathrm{in}},\, \finp_{n_\mathrm{in}+1},\,...\,,\,\finp_{n_\mathrm{in}+n_\mathrm{out}},\, ...\,,\,\finp_{n_\mathrm{in}+k n_\mathrm{out}}\right] 
\ee
are divided into $k+1$ sequences of length $n_\mathrm{in}$ each, which are shifted by $i = \{0\, n_\mathrm{out},\, ... , \,k \,n_\mathrm{out}\}$ time-steps. Each sequence is passed through the neural network and the results are concatenated, see \fig{fig:OsciRecursivelyEst}. The error estimator predicts errors $\hat{e}$. 
In output segment 3, the input is set to zero, which was not part of the training data for the S-NN, but nevertheless handled  well. In contrast, in segment 5, the input amplitude is doubled, another input not part of the training data. There the S-NNs estimation maintains the phase well, but the error increases significantly, which is captured well by the EE-N. In the transition to segment 6 both networks briefly struggle, but the S-NN regains accuracy as the initial disturbances decay.

\subsection{Planar flexible slider-crank}\label{sec:sliderCrankSetup}

\begin{figure}
\centering
\begin{minipage}{0.64\textwidth}
\includegraphics[width=\textwidth]{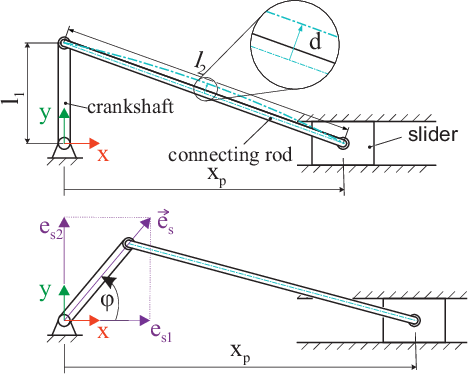}
\end{minipage}
\begin{minipage}{0.30\textwidth}
\begin{tabular}{c | c}
Parameter & value \\ \hline
$l_1$ & $\SI{0.5}{\metre}$ \\
$l_2$ & $\SI{1.0}{\metre}$ \\
$m_1$ & $\SI{2.0}{\kilogram}$ \\
$m_2$ & $\SI{4.0}{\kilogram}$\\
$m_3$ & $\SI{1.0}{\kilogram}$\\
$E I$ & $13.8 \si{\newton\metre\squared}$\\ 
$E\!A$ & $414 \cdot 10^3 \si{\newton}$\\ 
$d_{EI}$ & $0.015 E I$\\ 
$d_{EA}$ & $0.01 E A$\\ 
\end{tabular}	
\end{minipage}
\caption{
The slider-crank model. The connecting rod is flexible and has a deflection $d$. Body 1 is the crankshaft, body 2 the connecting rod, and 3 the slider. $EI$ describes the bending stiffness and $E\!A$ the axial stiffness of the connecting rod, whereas $d_{EI}$ and $d_{EA}$ are the bending and axial damping of the beam. Buckling is not considered by the model. The mass of the crankshaft 
is assumed to be distributed homogeneously, thus the inertia follows to $\frac{1}{12}m\,l^2$. 
}
\label{fig:models}%
\end{figure}

\alex{
    An idealized slider-crank mechanism composed from rigid bodies has only a single degree of freedom (DOF).
}
The slider's position $x_{p}$ can be calculated from the kinematics equation
\be
x_{p} = l_1 \cos(\varphi) + \sqrt{l_2^2 - l_1^2 \sin(\varphi)^2} \eqcomma
\ee
from the length of the crankshaft $l_1$, the length of the connecting rod $l_2$, and the angle $\varphi$ which is a possible minimal coordinate in the rigid mechanism.
 
\begin{figure}
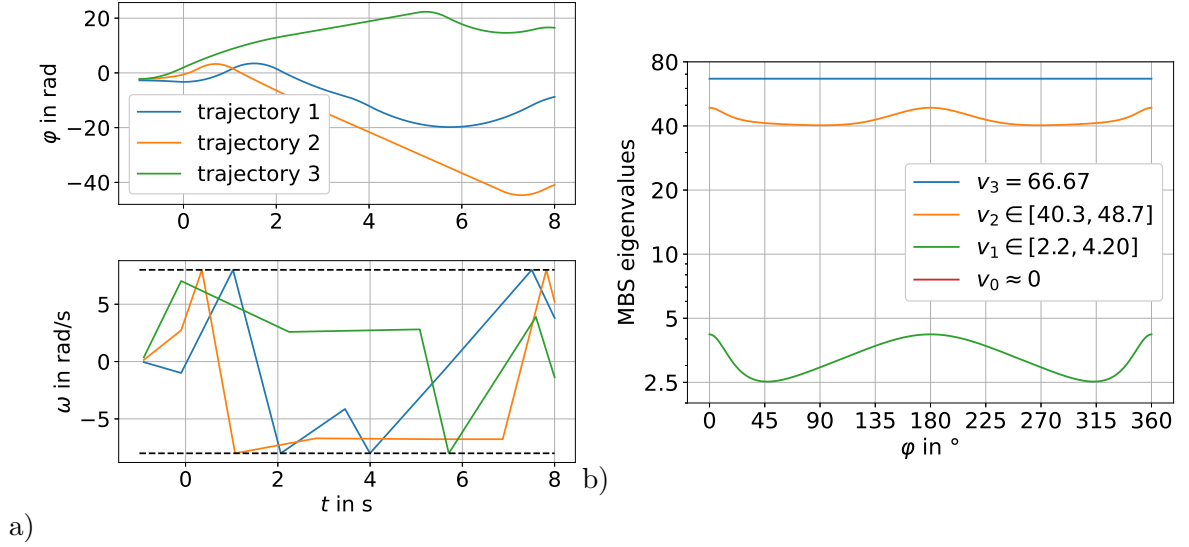
%
\centering
\minipageFig{0.46}{SliderCrankResults/Example_TrajectoryType0}{a)}
\minipageFig{0.48}{SliderCrankResults/EigenValues_V3}{b)}
\caption{a) Examples for applied trajectories to the crankshaft, where angular velocities change with constant acceleration. 
The angle $\varphi$ is initialized in the range of $\pm \pi$. 
b) The system's eigenvalues $v_1$ to $v_3$ plotted logarithmically over the angle of the crankshaft $\varphi$. 
}
\label{fig:TrajectoriesSC}
\end{figure}

In this example, a flexible slider-crank mechanism is modeled, in which the input is defined to be the desired angle $\varphi_\mathrm{des}$, for which trajectories with constant accelerations are created as shown in \fig{fig:TrajectoriesSC}. To follow the desired $\varphi$ and $\dot{\varphi}$, PD control is used and the torque
\be
\tau = P\, (\varphi_\mathrm{des} - \varphi)  + D (\dot{\varphi}_\mathrm{des} - \dot{\varphi}) \label{eq:SCControl}
\ee
is applied to the crankshaft. 
Slider and crankshaft are rigid bodies, whereas the connecting rod is modeled as a fully nonlinear Bernoulli-Euler beam following the absolute nodal coordinate formulation as described in~\cite{2008_gerstmayr_correctRepresentationANCF}. 
The ANCF beam is chosen for the flexible connecting rod in order to simplify reproducibility of results. 
The straightforward way to select the neural network's input would be the desired angles $\varphi_\mathrm{des}(t)$ or the desired angular velocities $\omega_\mathrm{des}(t)$ with the starting angle $\varphi_\mathrm{des}(0)$. An issue with the angle parametrization is the input normalization: if normalized in $[-\pi,\, \pi]$ there is a discontinuity at $\pm \pi$. Therefore, a parametrization using the director
\be
\ev_s = \begin{bmatrix}
e_\mathrm{s1} \\ e_\mathrm{s2}
\end{bmatrix} = 
\begin{bmatrix}
\cos{(\varphi_\mathrm{des})} \\ \sin{(\varphi_\mathrm{des})}
\end{bmatrix}
\ee
instead of the desired angle is chosen, which solves both the discontinuity and normalization problem. 
The neural network's input follows as
\be
\xv = \begin{bmatrix}\ev_{s \,0}^\mathrm{T}, \, \ev_{s\,1}^\mathrm{T}, \,..., \, \ev_{s \,(\nin-1)}^\mathrm{T}\end{bmatrix}^\mathrm{T} \eqcomma
\ee
with unit vectors $\ev_{s, i}$ in the $i$-th time-step of the dataset. 
To create training and validation data, the mechanical system is initialized at a randomized angle $\varphi_0$. 
\alex{
    As a consistent random initialization with arbitrary deflections and velocities is a non-trivial task,
}%
the angular velocity is initialized by adding a $\SI{1}{\second}$ startup phase, shown between times $\SI{-1}{\second}$ and $\SI{0}{\second}$, which is not part of the training data. 
The angular velocity undergoes constant acceleration, see \fig{fig:TrajectoriesSC}. 
Each acceleration phase takes between $20$ and $60$ time-steps and with a probability of $10\%$ the acceleration is set to zero. 
The black dashed line is the maximum velocity $\pm \left|\omega_\mathrm{max}\right| = \SI{8}{\radian\per\second}$, $\dot{\omega}_\mathrm{max} = \SI{20}{\radian\per\second\squared}$. 
The neural network learns both the dynamics of the mechanical system and the dynamics of the control from \eq{eq:SCControl}. 

The rigid body model has one degree of freedom, thereby, after eliminating the constraints, the eigenvalue $v_0 = 0$ corresponding to the rigid body motion persists. Also for the flexible slider crank model, the smallest eigenvalue is close to zero, thus practically undamped. 
The second smallest eigenvalue changes with the angle $\varphi$, which corresponds to \alex{decay times} of $\SI{1.1}{\second}$ to $\SI{1.83}{\second}$, shown in \fig{fig:TrajectoriesSC}. 
For a sequence duration of $\SI{4}{\second}$, input and output lengths of $n_\mathrm{in} = 128$ and $n_\mathrm{out} = 32$ are chosen. 

\begin{figure}%
\includegraphics[width=\columnwidth]{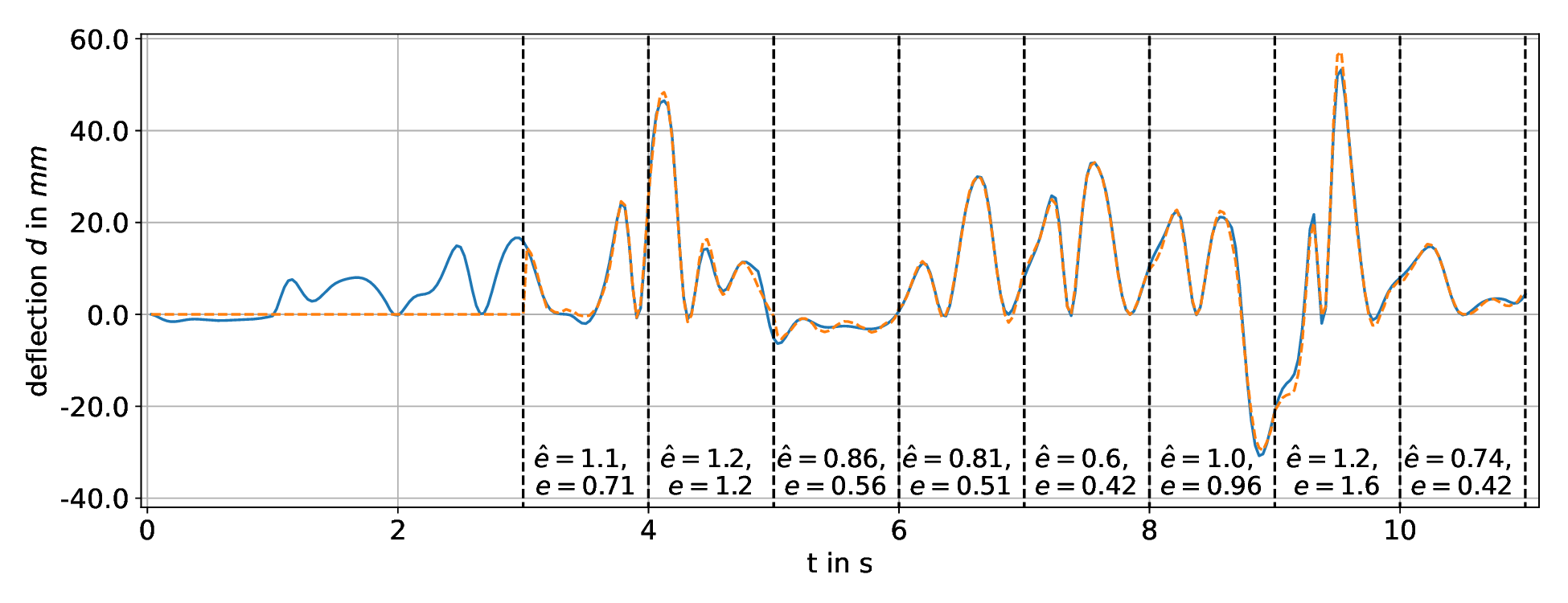}%
\caption{Processing a longer time-segment for the slider-crank system using the proposed method. For {$t < (n_\mathrm{in} - n_\mathrm{out})h$} no deflection can be estimated. The estimated error $\hat{e}$ and $e$ is shown for each segment in $mm$. } %
\label{fig:resultSCContEst}%
\end{figure}

In \fig{fig:resultSCContEst} the results of the SLIDE method is shown over a longer time period for the flexible slider crank system. This data is not part of the training or validation dataset. The orange dotted line is the surrogate model, while the blue solid line shows the ground truth obtained by the simulation model. The neural network is only trained to predict the last $n_\mathrm{out}$ steps of single sections and was not trained on continuation. The phase of the vibrations is well preserved over a longer period of time. The predicted error $\hat{e}$ and ground truth error $e$ are shown below.  


\subsection{Spatial 6R manipulator on a flexible socket} \label{sec:6R}
\begin{figure}%
\centering
\minipageFig{0.47}{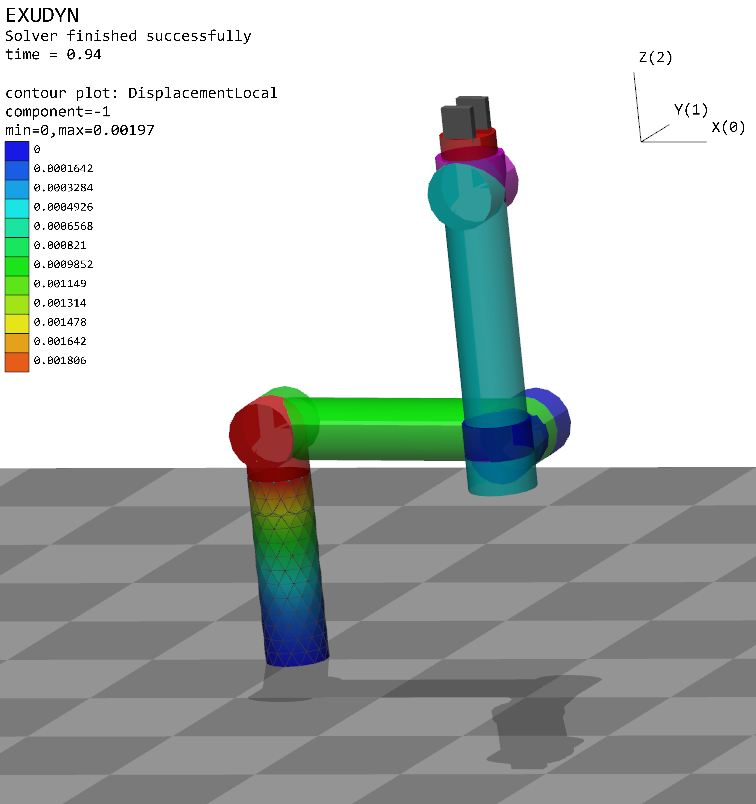}{a)}
\minipageFig{0.49}{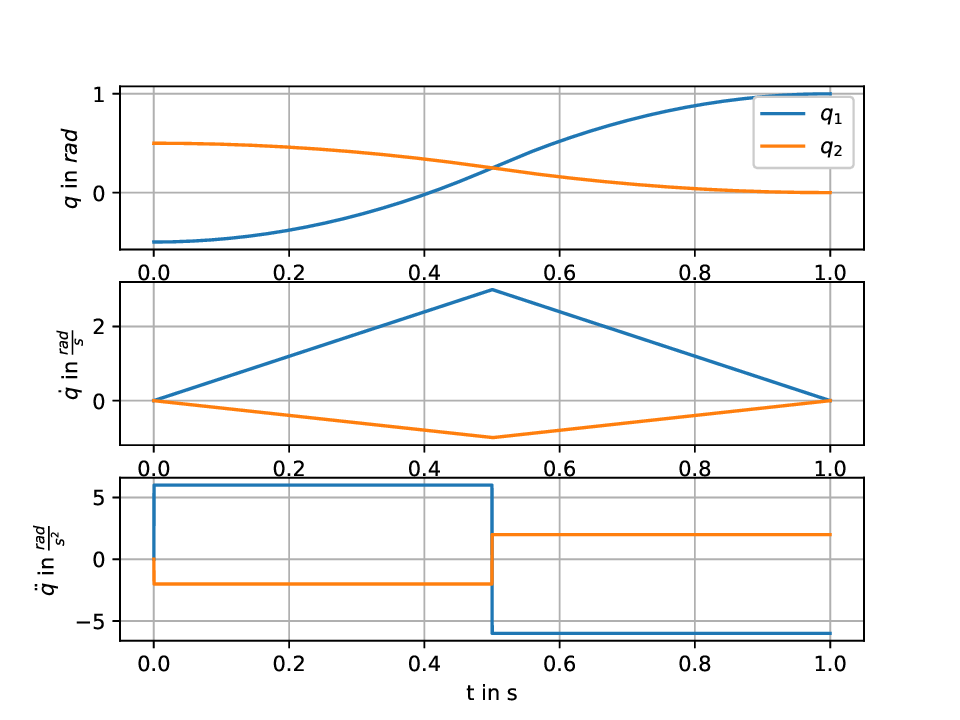}{b)}
\caption{
a) Simulation of the Puma560 manipulator standing on a flexible cylindrical socket.  
b) Example for a PTP motion for 2 joints. Between the start- and endpoint constant acceleration is applied, leading to linear changes in velocity and quadratic change in angles. 
}%
\label{fig:6RExample}%
\end{figure}

In this experiment the robotic manipulator puma 560 stands on a flexible socket, which deforms due to the forces induced by the robot's motion. 
The robot's mechanical parameters are taken from the robotics toolbox~\cite{2021_Corke_roboticstoolboxPython}\peter{, with minor adaptions. When the robot is mounted rigidly and the mounting point is static, the inertia around $x$ and $y$ of the first link, as well as the mass, are not part of the equations of motion~\cite{1986_armstrong_explicitDynamicModel}. To avoid unphysical behavior, we adjusted the first link's mass to $\SI{20}{\kilogram}$ and the missing inertias to $\SI{0.2}{\kilogram\metre\squared}$ to fulfill the triangular inequality. Furthermore the inertia of link 3 around its $z$-axis was doubled to $\SI{0.025}{\kilogram\metre\squared}$, also to fulfill the triangular inequality. }
The joint vector $\qv = \left[q_1, \, ... \,, q_6\right]$ starts at $\qv(t=0) = \qv_0$ and moves to $\qv (t=(n_{in}-n_{out})h) = \qv_1$ and to $\qv(t=n_{in} h) = \qv_2$ with a point-to-point (PTP) motion, applying constant acceleration to the joints. In \fig{fig:6RExample}, a) the simulation model of the robot is shown. 

The pose of the robot's tool center point (TCP) relative to the global frame can be described by the homogeneous transformation
\be
^{\mathrm{0, TCP}}\Tm(\qv(t)) = \LU{\mathrm{0, f}}{\Tm} \LU{\mathrm{f, TCP}}{\Tm} = 
\begin{bmatrix}
\LU{\mathrm{0, TCP}}{\Rm(\qv(t))} & \LU{\mathrm{0, TCP}}{\tv(\qv(t))} \\
\mathbf{0} & 1
\end{bmatrix}
\ee
with the rotation matrix $\Rm$ and translation vector $\tv$. 
Superscripts ${\mathrm{a, b}}$ of $\LU{\mathrm{a,b}}{\Tm}$ indicate that the pose of b is described relative to a. 
The robot's forward kinematics $\LU{\mathrm{f, TCP}}{\Tm}$ does not depend on the socket's deflection and can be calculated for given joint angles $\qv$. 
For a rigid socket $^\mathrm{0,f}\Tm$ is the pose of the robot's mounting point on the socket.
For the flexible socket the transformation to the flange $\LU{\mathrm{0,f}}{\Tm}$ changes with the displacement of the mesh. 
The error of the TCP position resulting from the flexibility results therefore in 
\be
\pv_\mathrm{e} = \LU{\mathrm{0,TCP}}{\tv} - \LU{\mathrm{f}, 0}{\Rm} (\LU{\mathrm{0,TCP}}{\tv} - \LU{\mathrm{0,f}}{\tv}) \eqdot
\ee
while the deviation of the rotation from the rigid solution is
\be
\Rm_{e} = \LU{\mathrm{EE}, 0}{\Rm} \LU{\mathrm{f, EE}}{\Rm}\eqdot
\ee
As the rigid body solution can be efficiently calculated by the forward kinematics, the neural network is trained on the deviations from the forward kinematics solution $\pv_e$. 
The robot is controlled by applying the torque $\tau_i$ to the $i$-th joint, calculated using PD control 
\be
\tau_i = P (q_{d\,i} - q_i) - D (\dot{q}_{d\,i} - \dot{q}_i)
\ee 
with control parameters $P$ and $D$, the desired angles $q_{d\,i}$ and angular velocities $\dot{q}_{d\,i}$. 
In between the angles $\qv_0$, $\qv_1$ and $\qv_2$, PTP interpolation with constant acceleration, shown in \fig{fig:6RExample}b), is applied to $\qv_d$. 
As torques and masses are not provided to the neural network, and it directly learns the mapping from $\qv_d$ to $\pv_e$, control and dynamic parameters are learned implicitly from the data. The control values $P = [4 \cdot 10^4,\;4\cdot10^4,\; 4\cdot10^4,\; 100,\;100, \;10]$ and $D = [400,\;400,\;100,\;1,\;1,\;0.1]$ are chosen arbitrarily.  

The flexible socket is modeled as a hollow cylinder with radius of $\SI{0.05}{\metre}$, $\SI{0.01}{\metre}$ wall thickness and $\SI{0.3}{\metre}$ length, Young's modulus $E = \SI{1}{\giga\pascal}$, density $\rho = \SI{1000}{\kilogram\per\metre\cubed}$ and Poisson’s ratio $\nu = 0.3$, which is in the range of some plastics such as high-density polyethylene (HDPE)~\cite{1999_kalay_enhancementMechanicalPropertiesHDPE}. 
The computed decay times of the robotic manipulator are shown in \fig{fig:6REigen} and are shortly elaborated. 
As the decay of the system's initial conditions can be approximated over a period of $T = n\,h$ steps with 
\be
A_\mathrm{rel}(T) \approx e^{\mathrm{Re}(v_{1})h} \;e^{\mathrm{Re}(v_{2})h} \;\; ... \;\; e^{\mathrm{Re}(v_{n})h} = e^{\sum_{i=1}^{n}\mathrm {Re}(v_{i})h} =  e ^{\mathrm{Re}(\bar{v})T} \eqcomma \label{eq:eigenValuesSum}
\ee
\alex{minima} in $v$, which correspond to maxima of the decay times $t_d$, 
have only a minor impact on the global behavior as the mean eigenvalues over the sequence $\bar{v}$ is the determining value. While the trajectory shown in the figure locally exceeds $t_d = \SI{10}{\second}$, the mean decay time to reach $A_\mathrm{rel} = 0.01$ is $\SI{0.32}{\second}$, the longest mean decay time calculated over $50$ randomized trajectories is $\SI{0.84}{\second}$ and the average $\SI{0.37}{\second}$. Thus the output window's truncation is chosen to $\SI{1}{\second}$. 
\begin{figure}%
\centering
\minipageFig{0.52}{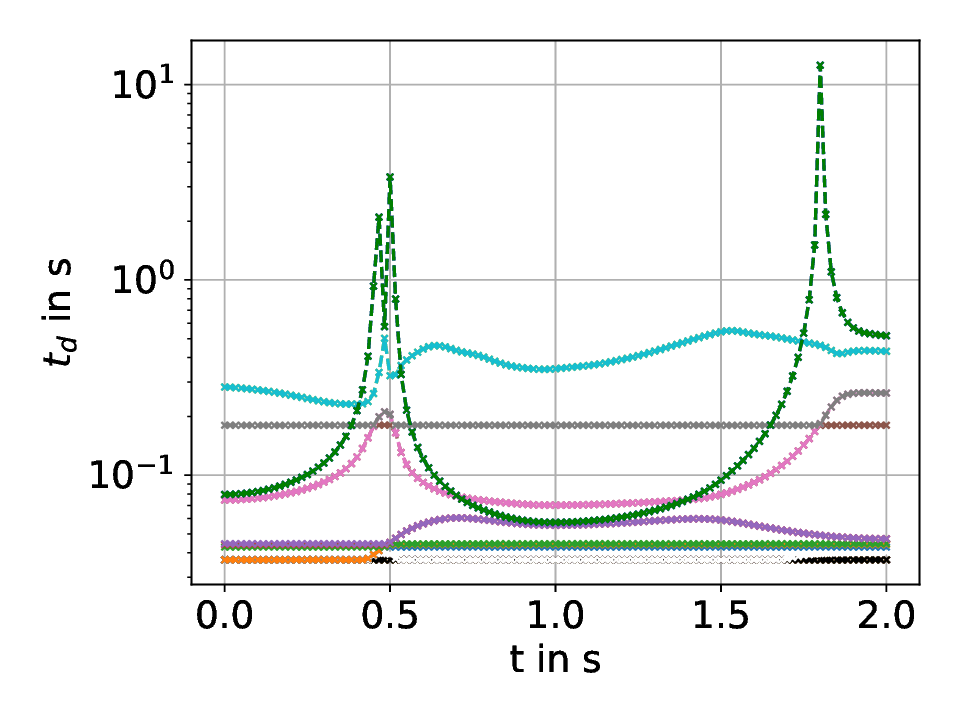}{}
\begin{minipage}[c]{0.3\textwidth}
\begin{tabular}{ c | c | c }
$\qv_0$ & $\qv_1$ & $\qv_2$ \\ \hline
 $-2.3$ & $1.83$ & $-2.38$ \\
 $-1.18$ & $-0.55$ & $1.87$ \\
 $-2.14$ & $-1.03$ & $1.28$ \\
 $-1.09$ & $0.27$ & $-1.87$ \\
 $-0.39$ & $-1.25$ & $-1.73$ \\
 $-1.42$ & $1.88$ & $1.96$ \\
\end{tabular}
\end{minipage}
\caption{Decay times based on  the eigenvalues of the systems, calculated over an exemplarily PTP trajectory from $\qv_0$ to $\qv_1$ and $\qv_2$. 
While the robot is in motion the real part of some eigenvalues approach $0$, thus $t_d$ increase. The values for the angles is shown at the right side. }
\label{fig:6REigen}%
\end{figure}

%
For the neural network's input from $\qv(t)$ a total number of $n_\mathrm{in}$ time steps are equidistantly sampled from the PTP trajectory in the time range $\left[0, t_2\right]$ and scaled by $\frac{\pi}{2}$ 
\be
\hat{\xv} = \frac{1}{\pi} \begin{bmatrix}
q_{1,0} &  q_{2,0} & q_{3,0} & q_{4,0} &  q_{5,0} & q_{6,0} \\
\vdots & \vdots & \vdots & \vdots & \vdots & \vdots \\
q_{1,n_{in}} &  q_{2,n_{in}} & q_{3,n_{in}} & q_{4,n_{in}} &  q_{5,n_{in}} & q_{6,n_{in}}
\end{bmatrix}
\ee
to obtain a normalized input to the neural network. With $n' = n_{in} - n_{out}$, the output of the simulation model is sampled in the time range $\left[t_1, t_2 \right]$
\be
\yv = \begin{bmatrix}
\LURU{}{\pv}{e, 1}{\mathrm{T}} \\
\vdots \\
\LURU{}{\pv}{e, n_\mathrm{out}}{\mathrm{T}} 
\end{bmatrix}
= 
\begin{bmatrix}
[x_{e, n'} & y_{e, n'} & z_{e, n'}] \\
\vdots & \vdots & \vdots \\
[x_{e, n_\mathrm{in}} & y_{e, n_\mathrm{in}} & z_{e, n_\mathrm{in}}] \\
\end{bmatrix}
\ee
and then normalized 
\be
\yv_\mathrm{s} = f_\mathrm{scale}(\yv) \eqdot
\ee
The input is comprised of $n_\mathrm{in} = 120$ time steps and the output of $n_\mathrm{out} = 60$ with step-size $h={1}/{60}\,\si{\second}$. 
Note that $\yv_s$ and $\hat{\yv}$ consists of $x$, $y$ and $z$ deflections which can be differently scaled. 
The root error estimator network is trained not on each deflection separately, but the mean of the Euclidean error. Therefore, the neural network's output $\hat{y}$ must be scaled back using $f_\mathrm{scale}^{-1}$ as shown in \fig{fig:StructureEstimator}. The mean error 
\be
e = \frac{1}{n_\mathrm{out}} 
\begin{bmatrix}
\Vert \yv_{s,n'} - f_\mathrm{scale}^{-1}(\hat{\yv}_{n'})\Vert_2 \\
\vdots \\
\Vert \yv_{s,n_\mathrm{in}} - f_\mathrm{scale}^{-1}(\hat{\yv}_{n_\mathrm{in}})\Vert_2 \\
\end{bmatrix}^\mathrm{T} 
\begin{bmatrix}
1 \\ \vdots \\ 1
\end{bmatrix}
\ee
is calculated from the mean euclidean distance between the model's output and the predicted deflection for each time step. This mean error is logarithmically scaled and subsequently used to train the error estimator. 

In \fig{fig:6RContinuation}, the application of SLIDE is shown \ref{sec:MethodAsymWindow}. The EE-N estimates the RMSE of all deflections with great accuracy. 
\begin{figure}%
\centering
\includegraphics[width=0.9\columnwidth]{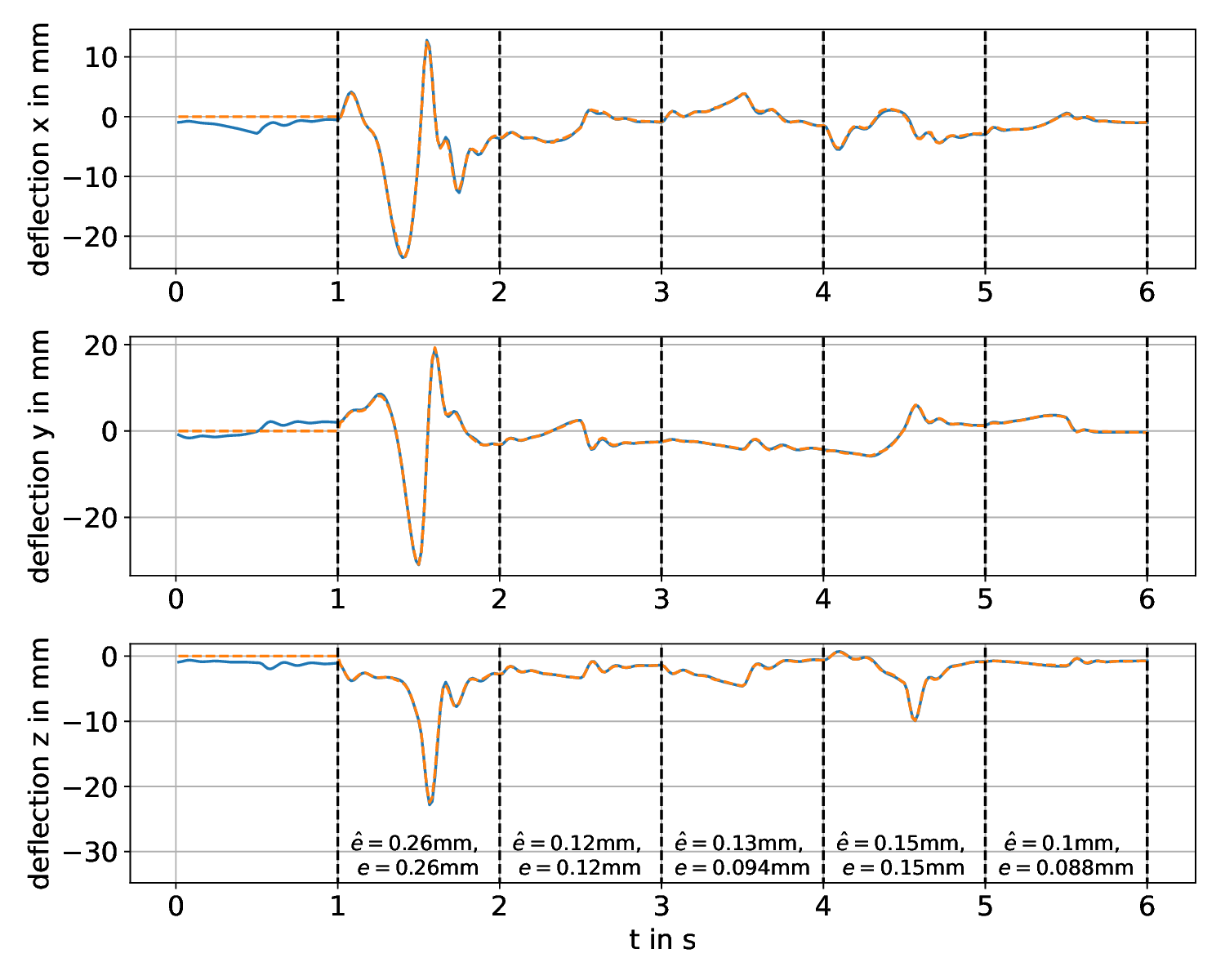}%
\caption{Continuation by window shifting with error of the surrogate model to the reference $e$ and estimated error $\hat{e}$, showing accurate approximation by the SLIDE method and reasonable estimation of the errors. }%
\label{fig:6RContinuation}%
\end{figure}
The desired angles for both control and model input are randomly sampled, and neither is part of the training or the validation set. In each output time segment $j$
\be
t^{(j)} = \left[t_{n' + j\,n_\mathrm{out}},\, t_{n' + j\,n_\mathrm{out} + 1},\, ...\,, \, t_{n' + (j+1) n_\mathrm{out}}\right] \eqcomma
\ee
both the reference and estimated error are shown, where the error estimation works with a accuracy of $2 - 12\%$ relative to the reference error. The accuracy of the surrogate model relative to the simulation model in the shown segments $t^{(1)}$ to $t^{(5)}$ is $\left[2.23,\, 2.44,\, 2.94,\, 3.45,\, 4.55\right] \%$. 

\begin{figure}%
\centering
\includegraphics[width=0.8\columnwidth, trim={0 0.3cm 0 1.8cm},clip]{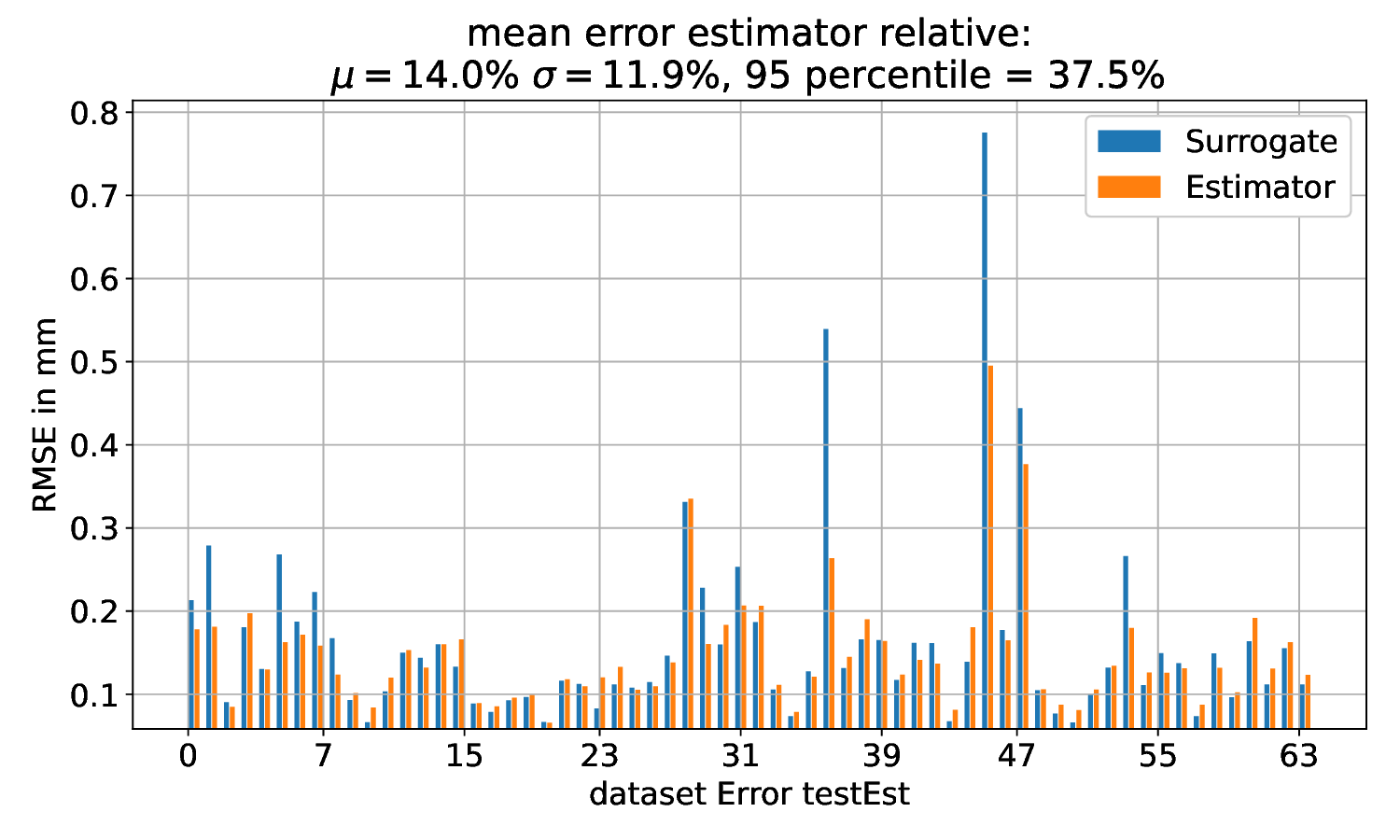}%
\caption{The mean absolute error of the error estimator and the surrogate model on parts of the test set of the flexible manipulator. }%
\label{fig:6REstimator}%
\end{figure}

\begin{figure}[tb]%
\centering
\includegraphics[width=0.8\columnwidth]{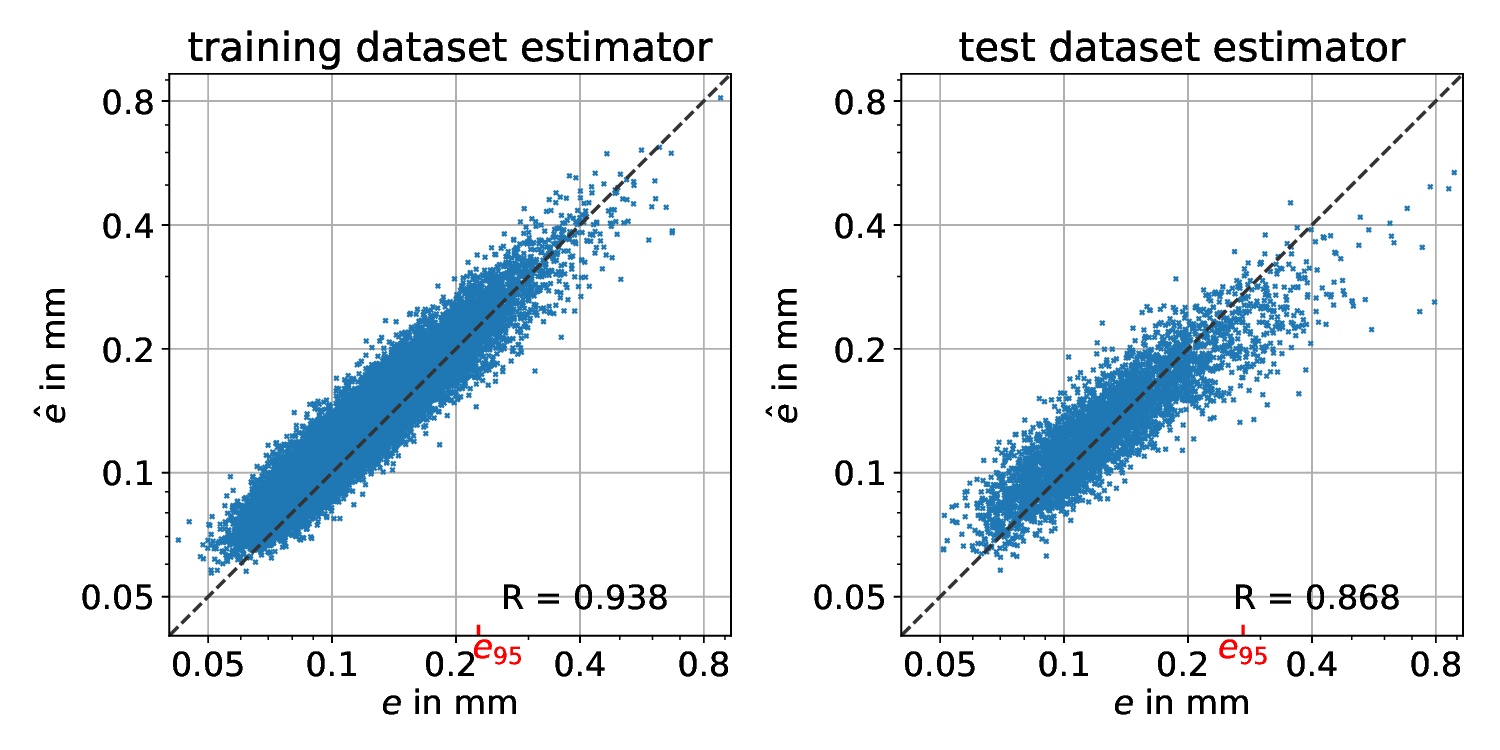}%
\caption{The error of the surrogate model $e$ is plotted over the estimated error $\hat{e}$, where $e_{95}$ marks the 95th percentile of the surrogate model error. Thus, for the training, the surrogate model has an accuracy of $e_{95}=\SI{0.227}{\milli\metre}$ or better on $95\%$ of the dataset. In the test $e_{95}= \SI{0.273}{\milli\metre}$. In the estimator's training set the surrogate training set is contained. }%
\label{fig:6RCorrelation}%
\end{figure}

For better understanding of the accuracy, the mean absolute error of the first 64 data points from the estimator's validation set are shown in \fig{fig:6REstimator}. The estimated error $e$ of the surrogate model is shown in blue and compared with the estimated $\hat{e}$. The error estimator predicts the mean absolute error of the surrogate model in the mean over the whole test set with $14.0\%$ and a standard deviation of $11.9\%$, while $95\%$ of the trajectories are within $37.5\%$ of the estimated error. 
The correlation of the surrogate errors and estimated errors is shown in \fig{fig:6RCorrelation}. The training dataset of the estimator is partly composed of the training dataset of the surrogate model. 


\subsection{Summary of results}
In the following section, the computational results for all experiments are summarized. 
As a general observation, while more neurons generally can represent a more complex behavior, networks with less neurons than the number of independent inputs (or outputs) require a higher compression of information and thus may lower the accuracy. In case of the Duffing oscillator, using less than $n_\mathrm{in}$ neurons decreases performance. 


\begin{table}
\begin{tabular}{l | c | c | c | c}
 & spring-damper & Duffing oscillator &  slider-crank & 6R robot $^{*}$ \\ \hline \hline
simulation 
& $\SI{7.30}{\milli\second} \pm \SI{103}{\micro\second}$ 
& $\SI{18.4}{\milli\second} \pm \SI{264}{\micro\second}$
& $\SI{748}{\milli\second} \pm \SI{12.1}{\milli\second}$
& $\SI{4.76}{\second} \pm \SI{30.5}{\milli\second}$\\ \hline
NN training 
& $\SI{14.8}{\second} \pm \SI{0.71}{\second}$ 
& $\SI{35.3}{\second} \pm \SI{0.13}{\second}$
&$\SI{549}{\second} \pm \SI{0.44}{\second}$ 
& $\SI{641}{\second} \pm \SI{1.18}{\second}$\\ \hline 
S-NN forward pass 
& $\SI{43.7}{\micro\second} \pm \SI{717}{\nano\second}$
& $\SI{126}{\micro\second} \pm \SI{259}{\nano\second}$ 
& $\SI{246}{\micro\second} \pm \SI{9.07}{\micro\second}$ 
& $\SI{625}{\micro\second} \pm \SI{22.9}{\micro\second}$\\ \hline
\shortstack[l]{validation set\\ mean RMSE}& $3.12 \cdot 10^{-7}$ & $6.45 \cdot  10^{-3}$ & $0.024$ & $0.020$ \\ \hline
speedup & 146 & 46.7 & 795 & 3247 
\end{tabular}
\caption{Numerical results on simulation and neural network (NN) training and forward pass times and accuracy. The RMSE is on the training of the S-NN. The speedup is depending on the length of the sliding windows as described in \eq{eq:speedup}. No batching is used. $^{*}$The training for the slider-crank and 6R robot model is run using a GPU. In training, this yields a speedup of $\approx 5.9$.}
\label{tab:allResults}
\end{table}

In \mytab{tab:allResults} the results of the experiments are shown, including time durations recorded by Python's internal \texttt{timeit} functionality. 
While the simulation runs for $n_\mathrm{in}$ steps, the neural network forward pass only obtains $n_\mathrm{out}$ steps. 
Therefore, the speedup $S$ follows to
\be
S = \frac{t_\mathrm{sim}}{t_\mathrm{NN}} \frac{n_{out}}{n_\mathrm{in}} \eqdot \label{eq:speedup}
\ee
Note that the values can be tweaked depending on the requirement for the application. In general, by increasing the amount of training data, the required training time increases and the error on the validation set decreases up to a certain point. Using larger and/or deeper neural networks increases times for both training and forward passes, but can simultaneously decrease the validation set RMSE, although more training data may be required to avoid overfitting. 

For the shown computations an 
i5-13400F CPU with up to $\SI{4.6}{\giga\hertz}$ and a Nvidia RTX 4070 
Graphics Processing Unit (GPU) were used. For both the spring-damper and slider-crank example, no significant speedup when using the GPU was visible in training or the forward pass -- supposedly the neural network and data size is not large enough. For the robot on the flexible socket, the speedup due to parallelization on the GPU is in the order of $6$ for the training. 
It should be highlighted that batching, similar to the batch size in the training process, increases the performance on the GPU significantly in the explored examples, because it allows the hardware to be better utilized.

\begin{figure}[tb]
\centering
\includegraphics[width=0.7\columnwidth]{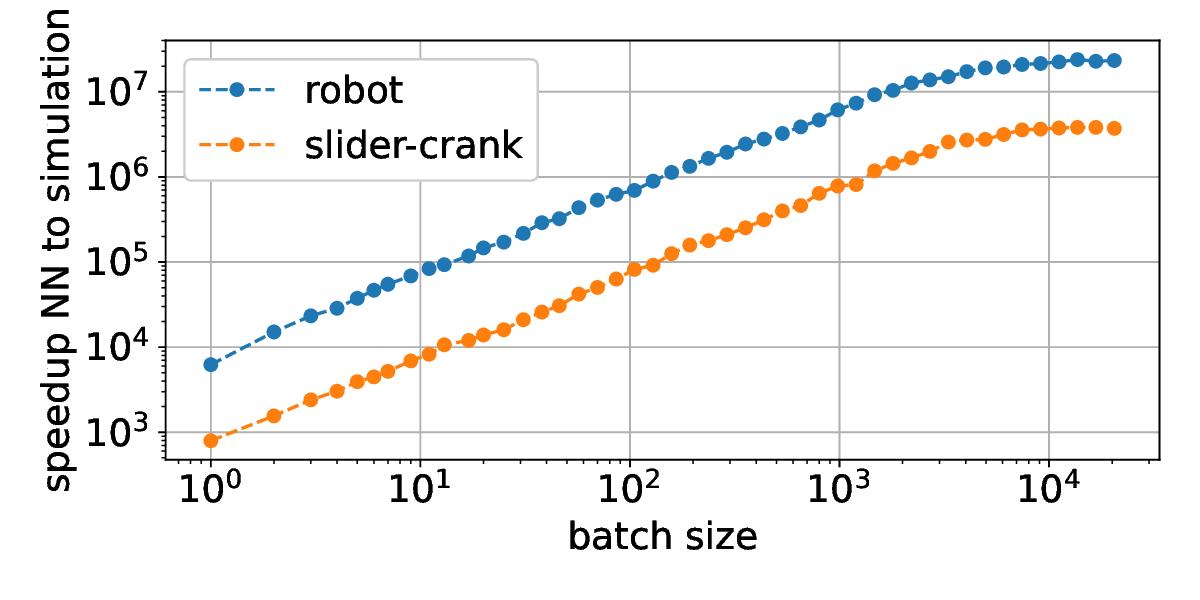}%
\caption{Speedup of neural network compared to multibody simulation of the shown manipulator for increasing batch size. The simulation model of the slider-crank is faster, thus less speedup is achieved. }%
\label{fig:SpeedupBatchsize}%
\end{figure}

Similar to the training process, where the input is not traversed one dataset at a time, but multiple datasets are processed at once, the input can be batched also in the analysis of longer time sequences or parallel simulations. For the data shown here, which is small compared to many other deep learning applications, increasing the number of batches only marginally impacts the computation time: a single forward pass for the trained network on the 6R example, shown in Appendix \ref{sec:Appendix}, 
takes $\SI{747}{\micro\second}$, whereas $1200$ batched trajectories take only $\SI{776}{\micro\second}$. The speedup resulting from increasing the batchsize is shown in \fig{fig:SpeedupBatchsize}. 
Through the SLIDE method, the input can also be segmented into batches, therefore a simulation of $\SI{1200}{\second}$ can be run in less than a millisecond, yielding a speedup of $7.36 \cdot 10^{6}$ compared with $\SI{2.38}{\second}$ of CPU-time per second of simulation, which enables real-time ``simulation'' of flexible bodies with ease. 
The maximum speedup for a simulation with a time-span of $\SI{13657}{\second}$ has been observed as $23.9 \cdot 10^6$, i.e., the simulation would run 9 hours, while the SLIDE method requires only $\SI{1.3}{\ms}$ for the prediction of this time span.

%% file: 05_Conclusion.tex
\section{Conclusions}
In this paper, we introduced SLIDE, the SLiding-window Initially-truncated Dynamic-response Estimator, 
a new method that leverages the computational power of GPUs to estimate dynamic system responses. 
We demonstrated its effectiveness on both classical mechanical systems, such as the Duffing Oscillator, and complex multibody systems, including a slider-crank and a 6R manipulator mounted on a flexible socket.

The only requirement for application of the SLIDE method is that the dynamic response is only affected by a (short) history of inputs, which is within the sliding input window. 
As only restriction, the examined system may not contain hidden internal states or effects like bifurcation, plasticity, or stick-slip effects, which are not contained in the SLIDE method's inputs but affect the output beyond the truncated window. 
By truncating the output window, SLIDE eliminates the need to (exactly) know dissipating initial conditions, which is especially beneficial for flexible multibody systems, where measuring the flexible coordinates is often unpractical. We also presented a practical approach for estimating decay times in multibody systems by linearizing the equations of motion, 
and calculating the eigenvalues. 
Possible applications of the shown method include, but are not limited to, 
selecting in real-time an optimal trajectory, while taking into account positioning errors due to deflection. 

In contrast to other approaches like Hamiltonian and Lagrangian neural networks~\cite{2019_Greydanus_hamiltonianNN, 2020_Cranmer_lagrangianNN}, the SLIDE method can be directly applied to non-autonomous systems with actuation and -- although we applied it to the field of multibody dynamics -- it could be used in other fields of simulation easily, as there are no assumptions on the structure of equations. 
%
For processing time sequences, recurrent neural networks (RNNs) or Long Short-Term Memory (LSTM)~\cite[chapter 10]{2016_Goodfellow_DeepLearning} are commonly used, as they are designed to process a sequence of values. 
Future research could focus on these approaches and implementations to provide a hidden state, supporting a broader class of systems. In addition, other architectures such as the Transformer~\cite{2017_Vaswani_attentionIsAllYouNeed} could be applied to dynamic problems and the error estimator to improve the accuracy.



%% file: Appendix.tex

\appendix
\section{Appendix: Neural network parameters}\label{sec:Appendix}
If not specified separately, the parameters from \mytab{tab:TrainingGeneral} are used. 
\peter{For convenience, \texttt{flatten} and \texttt{unflatten} is part of the sequential network to shape the data accordingly. Apart from the shown tests, residual connections and convolutional layers have been experimented with, but no significant improvement in the results was achieved. }
\begin{table}
\centering
 \begin{tabular}{ c | c | c | c  }
\hline
parameter & value & parameter & value \\ \hline
optimizer: & ADAM~\cite{2014_Kingma_AdamOptimizer} & variable type: & \texttt{float32} \\ \hline
learning rate: & $10^{-3}$ & batchsize & $n_{train}/8$ \\ \hline
size training set $n_{train}$&  1024 ... 20480 & size validation set $n_{val}$ & 64 ... 2048  \\ \hline
validation frequency & every 20 epochs & & \\ \hline
\hline
\end{tabular}
\caption{General parameters for training the neural network. For the ADAM optimizer the standard parameters from pytorch are used. 
}
\label{tab:TrainingGeneral}
\end{table}
\subsection{Duffing oscillator}
For the duffing oscillator shown in section~\ref{sec:DuffingOsc}, the size of the dataset is $4096$ for the training and $512$ for the validation. 
The neural network consists of two layers with $100$ neurons and ReLU activation function.  

\subsection{Flexible slider-crank}
The network in the flexible slider-crank example, see section~\ref{sec:sliderCrankSetup}, uses $6$ layers with ELU activation function and $192$ neurons each. The length of the input sequence is $n_\mathrm{in} = 128$
 time-steps and the output is $n_\mathrm{out} = 32$. The S-NN is trained for $2000$ steps and the EE-N for $500$ steps. Learning rate  is $1.5 \cdot 10^{-3}$.  

\subsection{6R-Robot on flexible socket}
For the results of section~\ref{sec:6R}, the 6R manipulator on a flexible socket, 
	the neural network is divided into 3 sub-networks, where each is associated with $x$, $y$ and $z$. They share the input of size $720 = n_\mathrm{in} * 6$ to $240$, followed by three ELU activation functions and $180$ neurons. 
By dividing the output into three networks, the number of weights is reduced greatly in the dense layers: The dense layers in the subnetworks have a total of $n_\mathrm{l1} = 3 \cdot 180^2 + 180 = 97740$ parameters, whereas in one layer the number of parameters would be $n_\mathrm{l2} = (3 \cdot 180)^2 + 540 = 292140 \approx 3 n_\mathrm{l1}$. 
The error estimator has 360 neurons and 2 ReLU activation functions. 
The neural network's training set consists of $20480$ trajectories and the validation set includes $4096$ trajectories. Each dataset consists of $n_\mathrm{in} = 120$ and $n_\mathrm{out} = 60$ time-steps with an input time of $\SI{2}{\second}$. The learning rate is increased to $1.5 \cdot 10^{-3}$ and the S-NN is trained for 2000 epochs and EE-N for 800 respectively. The error mapping $\epsilon_+ = -1.5$ and $\epsilon_- = -4.5$. 


%% file: SLIDE_DeepLearningForMBS.bbl
\begin{thebibliography}{10}

\bibitem{1989_Hornik_FFNareUniversalFunApproximators}
K.~Hornik, M.~Stinchcombe, and H.~White, ``Multilayer feedforward networks are
  universal approximators,'' {\em Neural networks}, vol.~2, no.~5,
  pp.~359--366, 1989.

\bibitem{2012_krizhevsky_AlexnetPaper}
A.~Krizhevsky, I.~Sutskever, and G.~E. Hinton, ``Imagenet classification with
  deep convolutional neural networks,'' {\em Advances in neural information
  processing systems}, vol.~25, 2012.

\bibitem{2016_He_deepResidualLearningForImageRecognition}
K.~He, X.~Zhang, S.~Ren, and J.~Sun, ``Deep residual learning for image
  recognition,'' in {\em Proceedings of the IEEE conference on computer vision
  and pattern recognition}, pp.~770--778, 2016.

\bibitem{2013_Mnih_PlayingAtari_DQN}
V.~Mnih, K.~Kavukcuoglu, D.~Silver, A.~Graves, I.~Antonoglou, D.~Wierstra, and
  M.~Riedmiller, ``Playing atari with deep reinforcement learning,'' {\em arXiv
  preprint arXiv:1312.5602}, 2013.

\bibitem{2016_Silver_masteringGoWithNN}
D.~Silver, A.~Huang, C.~J. Maddison, A.~Guez, L.~Sifre, G.~Van Den~Driessche,
  J.~Schrittwieser, I.~Antonoglou, V.~Panneershelvam, M.~Lanctot, {\em et~al.},
  ``Mastering the game of go with deep neural networks and tree search,'' {\em
  nature}, vol.~529, no.~7587, pp.~484--489, 2016.

\bibitem{2017_Vaswani_attentionIsAllYouNeed}
A.~Vaswani, N.~Shazeer, N.~Parmar, J.~Uszkoreit, L.~Jones, A.~N. Gomez,
  {\L}.~Kaiser, and I.~Polosukhin, ``Attention is all you need,'' {\em Advances
  in neural information processing systems}, vol.~30, 2017.

\bibitem{2022_VuQuocHumer_DeepLearningComputationalMechanics}
L.~Vu-Quoc and A.~Humer, ``Deep learning applied to computational mechanics: A
  comprehensive review, state of the art, and the classics,'' {\em arXiv
  preprint arXiv:2212.08989}, 2022.

\bibitem{2023_rabczuk_machineLearningInModelingAndSimulation}
T.~Rabczuk and K.-J. Bathe, ``Machine learning in modeling and simulation,''
  {\em Springer Cham, Switzerland}, vol.~10, pp.~978--3, 2023.

\bibitem{2019_Raissi_PINNs}
M.~Raissi, P.~Perdikaris, and G.~E. Karniadakis, ``Physics-informed neural
  networks: A deep learning framework for solving forward and inverse problems
  involving nonlinear partial differential equations,'' {\em Journal of
  Computational physics}, vol.~378, pp.~686--707, 2019.

\bibitem{2019_Greydanus_hamiltonianNN}
S.~Greydanus, M.~Dzamba, and J.~Yosinski, ``Hamiltonian neural networks,'' {\em
  Advances in neural information processing systems}, vol.~32, 2019.

\bibitem{2020_Cranmer_lagrangianNN}
M.~Cranmer, S.~Greydanus, S.~Hoyer, P.~Battaglia, D.~Spergel, and S.~Ho,
  ``Lagrangian neural networks,'' {\em arXiv preprint arXiv:2003.04630}, 2020.

\bibitem{2021_Cai_PINNsForFluidDynamics}
S.~Cai, Z.~Mao, Z.~Wang, M.~Yin, and G.~E. Karniadakis, ``Physics-informed
  neural networks ({PINN}s) for fluid mechanics: A review,'' {\em Acta
  Mechanica Sinica}, vol.~37, no.~12, pp.~1727--1738, 2021.

\bibitem{2021_Cai_PINNsHeatTransfer}
S.~Cai, Z.~Wang, S.~Wang, P.~Perdikaris, and G.~E. Karniadakis,
  ``Physics-informed neural networks for heat transfer problems,'' {\em Journal
  of Heat Transfer}, vol.~143, no.~6, p.~060801, 2021.

\bibitem{2021_Choi_dataDrivenSimulation_DeepNerualNetworks}
H.-S. Choi, J.~An, S.~Han, J.-G. Kim, J.-Y. Jung, J.~Choi, G.~Orzechowski,
  A.~Mikkola, and J.~H. Choi, ``Data-driven simulation for general-purpose
  multibody dynamics using deep neural networks,'' {\em Multibody System
  Dynamics}, vol.~51, pp.~419--454, 2021.

\bibitem{2021_Han_DNNFlexible}
S.~Han, H.-S. Choi, J.~Choi, J.~H. Choi, and J.-G. Kim, ``A {DNN}-based
  data-driven modeling employing coarse sample data for real-time flexible
  multibody dynamics simulations,'' {\em Computer Methods in Applied Mechanics
  and Engineering}, vol.~373, p.~113480, 2021.

\bibitem{2024_Pikulinski_DataDrivenInverseDynamics}
M.~Pikuli{\'n}ski, P.~Malczyk, and R.~Aarts, ``Data-driven inverse dynamics
  modeling using neural-networks and regression-based techniques,'' {\em
  Multibody System Dynamics}, pp.~1--26, 2024.

\bibitem{2021_Angeli_DeepLearningMinimalCoordinates}
A.~Angeli, W.~Desmet, and F.~Naets, ``Deep learning for model order reduction
  of multibody systems to minimal coordinates,'' {\em Computer Methods in
  Applied Mechanics and Engineering}, vol.~373, p.~113517, 2021.

\bibitem{2024_Slimak_overviewDesignConsiderationDataDrivenTimeStepping}
T.~Slimak, A.~Zw{\"o}lfer, B.~Todorov, and D.~J. Rixen, ``Overview of design
  considerations for data-driven time-stepping schemes applied to nonlinear
  mechanical systems,'' {\em Journal of Computational and Nonlinear Dynamics},
  vol.~19, no.~7, 2024.

\bibitem{2024_NajeraFlores_DaneQuinn_StrucurePreservingMLFrameworkIsolatedNonlinearities}
D.~A. Najera-Flores, D.~D. Quinn, A.~Garland, K.~Vlachas, E.~Chatzi, and M.~D.
  Todd, ``A structure-preserving machine learning framework for accurate
  prediction of structural dynamics for systems with isolated nonlinearities,''
  {\em Mechanical Systems and Signal Processing}, vol.~213, p.~111340, 2024.

\bibitem{2024_Gerstmayr_multibodyModelsFromNaturalLanguage}
J.~Gerstmayr, P.~Manzl, and M.~Pieber, ``Multibody models generated from
  natural language,'' {\em Multibody System Dynamics}, pp.~1--23, 2024.

\bibitem{2024_Ziegler_GitHubCopilotImpact}
A.~Ziegler, E.~Kalliamvakou, X.~A. Li, A.~Rice, D.~Rifkin, S.~Simister,
  G.~Sittampalam, and E.~Aftandilian, ``Measuring github copilot's impact on
  productivity,'' {\em Communications of the ACM}, vol.~67, no.~3, pp.~54--63,
  2024.

\bibitem{2024_Han_DataDrivenForcePredictionHydraulics}
S.~Han, G.~Orzechowski, J.-G. Kim, and A.~Mikkola, ``Data-driven friction force
  prediction model for hydraulic actuators using deep neural networks,'' {\em
  Mechanism and Machine Theory}, vol.~192, p.~105545, 2024.

\bibitem{2024_WangNegrut_MBDNode}
J.~Wang, S.~Wang, H.~M. Unjhawala, J.~Wu, and D.~Negrut, ``Mbd-node:
  Physics-informed data-driven modeling and simulation of constrained multibody
  systems,'' {\em Multibody System Dynamics}, pp.~1--43, 2024.

\bibitem{2015_Benner_survey}
P.~Benner, S.~Gugercin, and K.~Willcox, ``A survey of projection-based model
  reduction methods for parametric dynamical systems,'' {\em SIAM review},
  vol.~57, no.~4, pp.~483--531, 2015.

\bibitem{2020_Shabana_dynamicsofMBS}
A.~A. Shabana, {\em Dynamics of multibody systems}.
\newblock Cambridge university press, 5th~ed., 2020.

\bibitem{2024_Gerstmayr_Exudyn}
J.~Gerstmayr, ``Exudyn--a c++-based python package for flexible multibody
  systems,'' {\em Multibody System Dynamics}, vol.~60, no.~4, pp.~533--561,
  2024.

\bibitem{2016_Goodfellow_DeepLearning}
I.~Goodfellow, Y.~Bengio, and A.~Courville, {\em Deep Learning}.
\newblock MIT Press, 2016.

\bibitem{2022_Dubey_ActivationFunctionsSurveyAndBenchmark}
S.~R. Dubey, S.~K. Singh, and B.~B. Chaudhuri, ``Activation functions in deep
  learning: A comprehensive survey and benchmark,'' {\em Neurocomputing},
  vol.~503, pp.~92--108, 2022.

\bibitem{2020_Xu_HowNeuralNetworksExtrapolate}
K.~Xu, J.~Li, M.~Zhang, S.~S. Du, K.~Kawarabayashi, and S.~Jegelka, ``How
  neural networks extrapolate: From feedforward to graph neural networks,''
  {\em CoRR}, vol.~abs/2009.11848, 2020.

\bibitem{2010_Glorot_UnderstandingDifficultiesTrainingFFN}
X.~Glorot and Y.~Bengio, ``Understanding the difficulty of training deep
  feedforward neural networks,'' in {\em Proceedings of the thirteenth
  international conference on artificial intelligence and statistics},
  pp.~249--256, JMLR Workshop and Conference Proceedings, 2010.

\bibitem{2014_Kingma_AdamOptimizer}
D.~P. Kingma and J.~Ba, ``Adam: A method for stochastic optimization,'' {\em
  arXiv preprint arXiv:1412.6980}, 2014.

\bibitem{2024_Ansel_Pytorch2}
J.~Ansel, E.~Yang, H.~He, N.~Gimelshein, A.~Jain, M.~Voznesensky, B.~Bao,
  P.~Bell, D.~Berard, E.~Burovski, {\em et~al.}, ``Pytorch 2: Faster machine
  learning through dynamic python bytecode transformation and graph
  compilation,'' in {\em Proceedings of the 29th ACM International Conference
  on Architectural Support for Programming Languages and Operating Systems,
  Volume 2}, pp.~929--947, 2024.

\bibitem{2008_gerstmayr_correctRepresentationANCF}
J.~Gerstmayr and H.~Irschik, ``On the correct representation of bending and
  axial deformation in the absolute nodal coordinate formulation with an
  elastic line approach,'' {\em Journal of Sound and Vibration}, vol.~318,
  no.~3, pp.~461--487, 2008.

\bibitem{2021_Corke_roboticstoolboxPython}
P.~Corke and J.~Haviland, ``Not your grandmother’s toolbox--the robotics
  toolbox reinvented for python,'' in {\em 2021 IEEE international conference
  on robotics and automation (ICRA)}, pp.~11357--11363, IEEE, 2021.

\bibitem{1986_armstrong_explicitDynamicModel}
B.~Armstrong, O.~Khatib, and J.~Burdick, ``The explicit dynamic model and
  inertial parameters of the puma 560 arm,'' in {\em Proceedings. 1986 IEEE
  international conference on robotics and automation}, vol.~3, pp.~510--518,
  IEEE, 1986.

\bibitem{1999_kalay_enhancementMechanicalPropertiesHDPE}
G.~Kalay, R.~A. Sousa, R.~L. Reis, A.~M. Cunha, and M.~J. Bevis, ``The
  enhancement of the mechanical properties of a high-density polyethylene,''
  {\em Journal of applied polymer science}, vol.~73, no.~12, pp.~2473--2483,
  1999.

\end{thebibliography}
